\documentclass[11pt,a4paper]{article}
\usepackage{times,latexsym}
\usepackage{url}
\usepackage[T1]{fontenc}

\usepackage{booktabs}
\usepackage{graphicx}
\usepackage{amsmath}
\usepackage{amssymb}
\usepackage{bbm}
\usepackage{arydshln}
\DeclareMathOperator*{\argmax}{arg\,max}
\DeclareMathOperator*{\argmin}{arg\,min}
\usepackage[ruled,vlined,linesnumbered]{algorithm2e}
\usepackage{algpseudocode}
\usepackage{hyperref}
\usepackage{multirow}
\usepackage{colortbl}
\usepackage{bm}
\usepackage{soul}
\usepackage{adjustbox}
\usepackage{tcolorbox}
\usepackage{subcaption}
\usepackage{soul}
\tcbuselibrary{theorems,skins}
\usepackage{xcolor}
\definecolor{backgray}{rgb}{0.9,0.9,0.9}
\definecolor{textgray}{rgb}{0.6,0.6,0.6}
\definecolor{myblue}{HTML}{d4e1f5}
\definecolor{mygreen}{HTML}{CAE5CD}
\definecolor{myred}{HTML}{ffcccc}
\definecolor{eqblue}{HTML}{006EB8}
\definecolor{eqgreen}{HTML}{3C8031}
\definecolor{eqred}{HTML}{EE2967}
\definecolor{eqyellow}{HTML}{FFDF42}
\definecolor{eqorange}{HTML}{F26035}

\usepackage{annotate-equations}

\usepackage[acceptedWithA]{tacl2021v1}

\usepackage{xspace,mfirstuc,tabulary}

\newif\iftaclinstructions
\taclinstructionsfalse 
\iftaclinstructions
\renewcommand{\confidential}{}
\renewcommand{\anonsubtext}{(No author info supplied here, for consistency with
TACL-submission anonymization requirements)}
\newcommand{\instr}
\fi

\iftaclpubformat 

\else

\fi

\title{Not Eliminate but Aggregate: Post-Hoc Control over Mixture-of-Experts\\to Address Shortcut Shifts in Natural Language Understanding}

\author{
Ukyo Honda$^{1}$\qquad Tatsushi Oka$^{2}$\qquad Peinan Zhang$^{1}$\qquad Masato Mita$^{1}$\\
$^1$CyberAgent, Japan\qquad $^2$Keio University, Japan\\
\texttt{\{honda\_ukyo,zhang\_peinan,mita\_masato\}@cyberagent.co.jp}\\
\texttt{tatsushi.oka@keio.jp}
}

\date{}

\begin{document}
\maketitle
\begin{abstract}
  Recent models for natural language understanding are inclined to exploit simple patterns in datasets, commonly known as \emph{shortcuts}.
  These shortcuts hinge on spurious correlations between labels and latent features existing in the training data.
  At inference time, shortcut-dependent models are likely to generate erroneous predictions under distribution shifts, particularly when some latent features are no longer correlated with the labels.
  To avoid this, previous studies have trained models to eliminate the reliance on shortcuts.
  In this study, we explore a different direction: pessimistically aggregating the predictions of a mixture-of-experts, assuming each expert captures relatively different latent features.
  The experimental results demonstrate that our post-hoc control over the experts significantly enhances the model's robustness to the distribution shift in shortcuts.
  Besides, we show that our approach has some practical advantages.
  We also analyze our model and provide results to support the assumption.\footnote{The code is available at \url{https://github.com/CyberAgentAILab/posthoc-control-moe}.}
\end{abstract}

\begin{figure*}[t]
    \centering
    \includegraphics[width=0.9\textwidth,keepaspectratio]{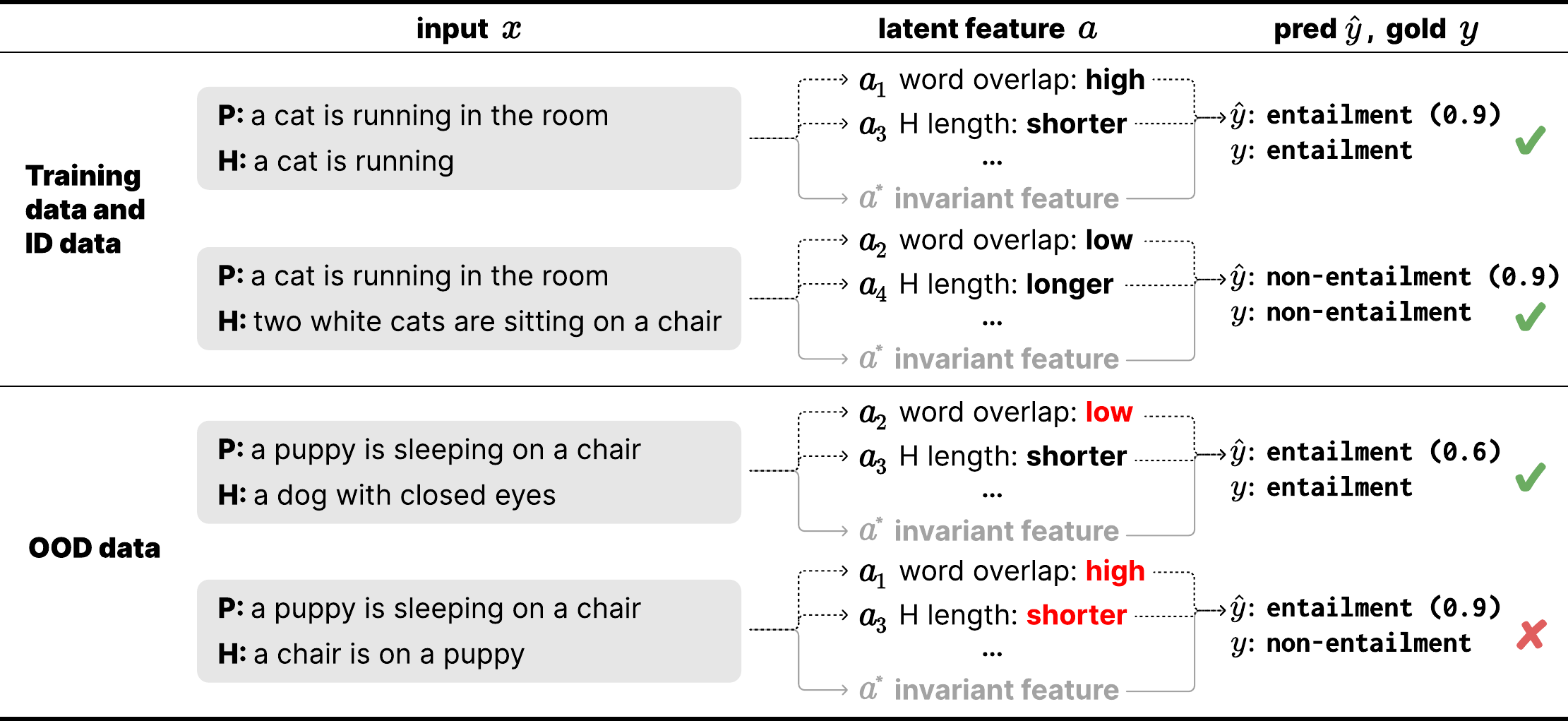}
    \caption{
    An illustrative example of shortcuts in the task of natural language inference.
    \emph{P} and \emph{H} denote the premise and hypothesis sentence, respectively.
    $\{a_i\}$ are latent features related to $x$.
    The value on the right-hand side of $\hat{y}$ shows the confidence $\in [0.0, 1.0]$ of the prediction.
    $a_i$ is correctly predictive of label $y$ in the training and ID data but not in the OOD data where the association between $a_i$ and $y$ changed.
    $a^*$ is an ideal latent feature predictive of $y$ across distributions.
    However, such $a^*$ is generally difficult for models to rely on. This figure illustrates the common case where \textcolor{textgray}{predictions are not based on $a^*$}.
    }
    \label{fig:overview}
\end{figure*}

\section{Introduction}
\label{sec:intro}
The datasets for natural language understanding (NLU) often contain simple patterns correlated with target labels, which are unintentionally introduced by annotators' simple heuristics, preferences, etc. \cite{gururangan-etal-2018-annotation,geva-etal-2019-modeling}.
More fundamentally, the compositional nature of natural language inherently introduces tokens that correlate with target labels individually \citep{gardner-etal-2021-competency}.
For example, word overlap \cite{mccoy-etal-2019-right,zhang-etal-2019-paws} and specific vocabulary, such as negations \cite{gururangan-etal-2018-annotation,schuster-etal-2019-towards}, are known to have such correlations.
However, these correlations are not guaranteed to hold in general and are therefore called \textbf{spurious correlations} \citep{feder-etal-2022-causal}.
The simple patterns are easy to exploit, so recent NLU models are inclined to take advantage of them.
This exploitation or the exploited patterns themselves are called \textbf{shortcuts} \citep{pmlr-v151-makar22a,feder-etal-2022-causal,du-etal-2021-towards,meissner-etal-2022-debiasing}.\footnote{Shortcuts are also called dataset \emph{bias}. However, we avoid using this term because it is confusing with the social bias or bias of an estimator. Similarly, we do not use the term \emph{debiasing} in this paper.}

At inference time, shortcuts often result in inaccurate predictions under relevant distribution shifts.
The shifts can occur, for example, when test data are collected from annotators with different heuristics or preferences \citep{geva-etal-2019-modeling,mccoy-etal-2019-right,zhang-etal-2019-paws,schuster-etal-2019-towards}.
Data from the same distribution as the training data is referred to as \textbf{in-distribution (ID)} data, while data from a distribution shifted relative to the training data is referred to as \textbf{out-of-distribution (OOD)} data.
Figure~\ref{fig:overview} shows the examples of ID and OOD data.

A simple solution to this problem is to eliminate reliance on shortcuts, which is the mainstream approach, including recent studies in NLU \citep{clark-etal-2019-dont,he-etal-2019-unlearn,mahabadi-etal-2020-end}.
Typically, those methods up-weight training instances where some known shortcuts cannot predict labels correctly and down-weight the others.
A practical deficiency of this approach arises in a performance trade-off between ID and OOD data.
It deviates models from ID data by eliminating shortcuts, which are valid features in ID data.
Due to this trade-off, another practical problem arises where the hyperparameter search has to be made using OOD test or validation data, as noticed as the limitation of previous work \citep{clark-etal-2019-dont,mahabadi-etal-2020-end,clark-etal-2020-learning,ghaddar-etal-2021-end,liu2021just,pmlr-v139-creager21a,yu-etal-2022-interventional,pmlr-v202-yang23s}.
Even when using OOD validation data, its distribution is the same as that of test data.
Thus, in other words, the approach requires knowing the test-time distribution to tune hyperparameters, which is impractical in testing OOD robustness.

In this paper, we opt not to pursue training to eliminate shortcuts.
Instead, we propose to aggregate predictions of a mixture model during inference.
The problem with shifts in shortcuts is that some latent features in the training data are no longer associated with the labels.
We hypothesize that this OOD situation can be addressed by effectively aggregating predictions, assuming that the predictions are based on relatively different latent features.
We propose a mixture model and its training strategy to encourage such modeling of latent features.
At inference time, we perform theoretically grounded risk minimization strategies through post-hoc control for the predictions in the event of potential shifts in shortcuts.

The experimental results demonstrate that our method significantly enhances the model's robustness when faced with shifts in shortcuts.
Moreover, our method shows two other practical benefits that address the problems of previous methods.
First, the mixture weights of our model can be used to detect shifts in latent features during inference.
This opens up the possibility of adaptive post-hoc control to address the performance trade-off between ID and OOD data.
Second, hyperparameters can be tuned with ID data only, removing the need to tune hyperparameters with OOD data.
We also analyze our mixture model and provide results supporting the assumption of modeling latent features.

\section{Background}
\label{sec:background}
This section first overviews shortcuts.
Then, we describe how previous approaches have addressed shortcuts and outline how we approach them.
Below, $\mathcal{X}$ and $\mathcal{Y}$ denote the input instance space and the entire class of target labels, respectively.

\subsection{Shortcuts in Detail}
\label{sec:shortcuts}
Shortcuts or spurious correlations arise when (1) some feature $a$ related to input $x \in \mathcal{X}$ is predictive of label $y \in \mathcal{Y}$ in training data, (2) but this association between $a$ and $y$ changes under relevant distribution shifts \citep{pmlr-v151-makar22a,feder-etal-2022-causal}.
Often, those features are latent, that is, difficult to identify \emph{a priori}.
Among those latent features, shortcuts refer to those that are easy to represent; sometimes, they refer to the exploitation of such latent features \citep{pmlr-v151-makar22a,feder-etal-2022-causal,du-etal-2021-towards,meissner-etal-2022-debiasing}.
Following \citet{pmlr-v151-makar22a}, we emphasize that the ease of modeling is an important characteristic of shortcuts.
It enables models to capture and depend on the latent features, thereby posing a serious threat when the relevant distribution shifts.

Shortcuts are pervasive in NLU datasets due to the simple heuristics, preferences, etc., possessed by annotators \citep{gururangan-etal-2018-annotation,geva-etal-2019-modeling}.
Shortcut-dependent models severely degrade performance on datasets collected with different heuristics and preferences \citep{geva-etal-2019-modeling,mccoy-etal-2019-right,zhang-etal-2019-paws,schuster-etal-2019-towards}.
Moreover, there is a more fundamental discussion that the compositional nature of natural language produces many simple features (\textit{e.g.}, words and phrases) that can robustly predict labels when the entire context is considered but are only spuriously correlated when considered individually \citep{gardner-etal-2021-competency,eisenstein-2022-informativeness}.
Figure~\ref{fig:overview} shows an illustrative example where simple word-overlap features and length features are associated with labels in training and ID data, but the association drastically changes in OOD data.

While pervasive, note that shortcuts are only part of the distribution-shift problem.
For example, shortcuts can be viewed as a special case of domain shift \citep{feder-etal-2022-causal} and can arise independently from the shift in label distribution $p(y)$ \citep{pmlr-v202-yang23s}.
Also, the problem of shortcuts is one of the consequences of underspecification, where distinct solutions can solve the problem equivalently well \citep{damour2022under}.
Following \citet{pmlr-v151-makar22a}, \textbf{we address distribution shifts exclusively in terms of shortcuts}.
Consequently, the OOD data we address involve shifts in the association between $a$ and $y$.

\subsection{Overview of Previous Approaches}
\label{sec:previous approaches}
To improve the OOD performance, previous studies have tried to remove the reliance on shortcuts.
In the study of NLU, a widely used approach is \textbf{reweighting} \citep{clark-etal-2019-dont,he-etal-2019-unlearn,mahabadi-etal-2020-end}.
This approach reweights instances to reduce learning on shortcut-inducing instances and increase learning on the others.
The weights are computed based on how accurately shortcuts predict labels.
During training, instances where shortcuts are predictive are down-weighted, and the others are up-weighted.

In the machine learning (ML) literature, training data is first partitioned into groups (also called environments) based on the spuriously correlated features.
The training data is assumed to be a mixture of the groups divided by the features.
Previous approaches avoid relying on shortcuts, that is, the group-specific spurious correlations.
There are two principal approaches in the ML literature.
Invariant risk minimization \citep[\textbf{IRM};][]{arjovsky2019invariant} trains a classifier that is simultaneously optimal for all groups.
Group distributionally robust optimization \citep[\textbf{GroupDRO};][]{sagawa2020distributionally} learns to minimize the worst-group risk by up-weighting the loss of the worst-case group.

\subsection{Problems of Previous Approaches}
\label{sec:problems}
These approaches share one common idea: training models while minimizing reliance on shortcuts to achieve robust predictions.
In practice, however, daring to eliminate predictive features in the training data and its ID data causes deviations from the ID data, resulting in a performance trade-off between the ID and OOD data.
See, for example, the aforementioned work on reweighting, IRM, and GroupDRO for empirical results.
This trade-off raises the following practical problems.

\paragraph{\colorbox{backgray}{(a)} Overfitting to OOD data.}
The degraded performance on ID data is a direct consequence of this trade-off \citep{utama-etal-2020-mind}.
Evaluating worst-case performance or performance on adversarial OOD data is essential for assessing generalization, and this study also aims to improve on these evaluations.
However, such extreme distribution shifts do not always occur after model deployment, so it is desirable for practical purposes to be able to deal with ID data as well.

\paragraph{\colorbox{backgray}{(b)} Hyperparameter tuning with OOD data.}
An indirect but more fundamental problem is the need for OOD test or validation data to tune hyperparameters.
This problem arises because the trade-off makes it difficult to predict performance on OOD data simply by looking at performance on ID data.
Obtaining OOD test data (in the context of ML, worst-group) requires pre-identification of shortcuts and their test-time distribution.
Even when using OOD validation data, its distribution needs to be the same as test data's.
In testing OOD robustness, this requirement is clearly impractical.

Initial studies used training data where shortcuts are pre-identified, in addition to OOD test or validation data \citep[][\textit{inter alia}]{clark-etal-2019-dont,arjovsky2019invariant,sagawa2020distributionally}.
Pre-identification of shortcuts is costly as it requires careful analysis of given data.
Seeking more practical solutions, subsequent approaches followed that did not require pre-identification of shortcuts in training data \citep[][\textit{inter alia}]{clark-etal-2020-learning,liu2021just,pmlr-v139-creager21a}.
However, they still need OOD test or validation data related to the pre-identified shortcuts to tune hyperparameters.
This requirement has been discussed as a serious limitation for practical use \citep{clark-etal-2019-dont,mahabadi-etal-2020-end,clark-etal-2020-learning,ghaddar-etal-2021-end,liu2021just,pmlr-v139-creager21a,yu-etal-2022-interventional}.
Moreover, the performance of those approaches on OOD data is considerably low without the hyperparameter tuning on OOD data \citep{pmlr-v202-yang23s}.

\subsection{Overview of Our Approach}
\label{sec:our approach}
In this study, we explore a different direction from previous approaches: We aggregate predictions of a mixture model.
Our hypothesis is that effective aggregation of predictions enables addressing potential shifts in shortcuts, suppose the predictions are based on relatively different latent features.

Let $a$ be a discrete random variable defined over the space $\mathcal{A} = \{a_1, ..., a_K\}$.
As described in Section~\ref{sec:shortcuts}, NLU data are likely to have multiple latent features associated with labels.
Considering the existence of those latent features, the conditional probability of $y$ given $x$, $p(y|x)$, can be rewritten as a finite mixture model as follows:\footnote{
We assume that it is reasonable to consider the finite latent features.
Generally speaking, model predictions are likely to depend on simple latent features strongly associated with labels, not evenly dependent on infinite possible latent features.
In addition, it is empirically established that finite mixture models can approximate a wide variety of distributions, as long as a sufficient number of mixture components are included \citep{titterington1985statistical,walker2011advances,nguyen2020approximation}.
}
\begin{align}
    p(y|x) &= \sum_{a \in \mathcal{A}} p(y, a|x), \\
    \label{eq:general mixture model}
    &= \sum_{a \in \mathcal{A}} \eqnmarkbox[eqgreen]{E}{p(y|a, x)} \eqnmarkbox[eqred]{R}{p(a|x)}.
\end{align}
\annotate[yshift=-0.6em]{below, left}{R}{mixture weights}\\
Therefore, the mixture model naturally aligns with the data structure.
The \emph{mixture weights} $p(a|x)$ are expected to estimate the distribution of the latent features, and $p(y|a, x)$ to predict based on each $a$.
If this assumption holds, pessimistic aggregation of $p(y|a, x)$ can minimize the risk under OOD circumstances at inference time, where it is not known which latent features to rely on.

In addition, this approach allows us to address the problems described in Section~\ref{sec:problems} as follows.
\colorbox{backgray}{\textbf{(a$'$)}} We have the flexibility to address both ID and OOD data scenarios through the adaptable application or omission of post-hoc control techniques.\footnote{While this adaptive use of post-hoc control necessitates determining whether the test data falls under the ID or OOD category, the results presented in Table~\ref{tab:router loss} suggest that changes in the mixture weights can effectively discern this distinction at inference time.
We revisit this point in Section~\ref{sec:analysis penalty}.
}
This adaptability sets our approach apart from the existing methods, which typically rely on fitting a single model exclusively for either the ID or OOD case.
\colorbox{backgray}{\textbf{(b$'$)}} Since our method focuses on fitting ID data during training, it does not require OOD data for training or tuning hyperparameters.

\section{Methods}
\label{sec:method}
The proposed method consists of two parts.
The first part is a training method using a mixture model.
The second part is a test-time operation, which aggregates the mixture model's predictions to make robust predictions when facing distribution shifts.
Figure~\ref{fig:method overview} in Appendix~\ref{sec:appx figures} shows the overview of our method.

\subsection{Training Phase: Mixture Model to Capture Latent Features}
\label{sec:mixture model}
To model latent features, we employ a mixture model, as seen in Eq~\eqref{eq:general mixture model}.
A typical implementation of the mixture model is \textbf{mixture-of-experts (MoE)} \citep{jacobs1991adaptive}.
\citet{shazeer2017outrageously} showed that MoE improves performance and efficiency in large-scale deep-learning models.
Following these studies' success, we employ a variant of MoE in this study.

\paragraph{Our Implementation of MoE.}
\label{sec:moe}
MoE consists of $K$ \emph{expert networks (experts)} and a \emph{router network (router)} responsible for assigning inputs to the expert networks.
Hereafter, we use the terms mixture weights and output distribution of a router interchangeably because they have the same function.
The MoE given an input $x$ is defined as follows:
\begin{align}
    \mathrm{MoE}(x) = 
    \sum^K_{k=1} \eqnmarkbox[eqgreen]{E}{E^k(x)} \eqnmarkbox[eqred]{R}{\pi_{k}(x)},
\end{align}
\annotate[yshift=0.6em]{above, right}{E}{$k$-th expert}
\annotate[yshift=-0.4em]{below, right}{R}{router}\\
where $E^k(x)$ is the output of the $k$-th expert and 
$\pi_{k}(x)$ is the $k$-th element of the router output 
$\pi(x) = \begin{bmatrix} \pi_{1}(x),\dots,\pi_{k}(x) \end{bmatrix}^{\top}\in \mathcal{P}$,
where $\mathcal{P}$ is a $(K-1)$-dimensional simplex.
The structure of the expert and router networks can be arbitrarily determined.
For example, the MoE modules can be stacked layer by layer while sparsifying the output of the router network \citep{shazeer2017outrageously}.

Our goal in employing the mixture model is to enable the aggregation of predictions based on relatively different latent features.
To this end, the parameters capturing latent features should be in one place.
Therefore, we decided to employ the MoE module only in the final layer of classification.
This is consistent with \textbf{mixture-of-softmax (MoS)}, a particular instantiation of MoE \citep{yang2018breaking}.
Similar to MoE, MoS consists of the experts and the router.
Both the expert and router receive the same encoded vector from an encoder.
Then,  the experts predict target labels, while the router determines the weights over the experts.
Our implementation of MoS is defined as follows:
\begin{align}
    \label{eq:MoS}
    p_{\bm{\theta}}(y|x) &= \sum^K_{k=1} \eqnmarkbox[eqgreen]{E}{p^{k}(y|x)} 
    \eqnmarkbox[eqred]{R}{\pi_{k}(x)},
\end{align}
\annotate[yshift=-0.4em]{below, right}{E}{$E^k$}\\
where we specify 
\begin{align}
    p^{k}(y| x) &= 
    \frac{
    \exp  \big (
    f^k(\mathbf{h})^{\top} \mathbf{w}^k_{y} + b^k_{y}
    \big)
    }{
    \sum_{y' \in \mathcal{Y}} \exp \big(f^k(\mathbf{h})^{\top} \mathbf{w}^k_{y'} + b^k_{y'}\big)
    }, \\
    \pi_{k}(x) &= 
    \frac{
    \exp  \big (f^r(\mathbf{h})^{\top} \mathbf{v}_k \big)
    }{
    \sum_{j=1}^{K} \exp \big(f^r(\mathbf{h})^{\top} \mathbf{v}_j \big)
    }, \\
    \label{eq:encoder h}
    \mathbf{h} &= g_{\bm{\phi}} (x).
\end{align}
Here, 
$g_{\bm{\phi}}: \mathcal{X} \rightarrow \mathbb{R}^d$
is the encoder and its $d$-dimensional output is denoted by $\mathbf{h}$.
$\mathbf{w}^k_{y}$ and $\mathbf{v}_k$
are the $d \times 1$ weighting vector
related to the $k$-th expert prediction for $y$
and the $k$-th element of the router output, respectively.
The functions
$f^k$ 
and 
$f^r$
respectively transform the encoder outcome
to $\mathbb{R}^d$
for the $k$-th expert and the router.
Both functions have the same structure and size of parameters, but the parameters are initialized and updated separately.
We employ BERT for $g_{\bm{\phi}}$ and set $f^{*}$ to the prediction head of BERT \citep{devlin-etal-2019-bert,zhang-etal-2022-making}: $f^*(\mathbf{h}) = \mathrm{LayerNorm} \circ \mathrm{ReLU} \circ \mathrm{Linear}(\mathbf{h})$, where $\circ$ represents composite functions that applied from right to left.
$\bm{\theta}$ denotes the entire parameters above.

We use the cross-entropy loss to train the parameters.
Given a mini-batch of $M$ instances with one-hot encoding of labels, the loss is as follows:
\begin{align}
    \label{eq:classifier loss}
    \mathcal{L}_C (\bm{\theta}) = - \frac{1}{M} \sum_{m=1}^{M} \log p_{\bm{\theta}} (y_m | x_m).
\end{align}

\paragraph{Penalty Term for $\pi$: Different Experts for Different Latent Features.}
\label{sec:router constraint}
Comparing Eq.~\eqref{eq:general mixture model} and \eqref{eq:MoS}, we see that different experts are expected to capture different latent features that predict labels.
However, this expectation does not hold when the mixture weights are consistently uniform or dominated by the same few experts across all the training instances.
In those cases, all the experts or the few experts capture the latent features indistinguishably.
To facilitate capturing the mixture of latent features at the mixture architecture, we propose a penalty term that constrains the router $\pi$.
Intuitively, it encourages the router to assign different inputs to different experts, assuming that different inputs have differences in their latent features to some extent.

Given a mini-batch of size $M$, define a $K \times M$ matrix of the router outputs as follows:
\begin{align}
    \mathbf{\Pi} &= 
    \begin{bmatrix}
        \pi(x_{1}), \pi(x_{2}), \dots, \pi(x_{M})
    \end{bmatrix}. 
\end{align}
Our goal is to encourage the columns of $\mathbf{\Pi}$ to be distinct distributions from each other.
Hinted by \citet{lin2017self}, we accomplish this by minimizing the Frobenius norm of 
$\mathbf{\Pi}^{\top} \mathbf{\Pi}$, where
the $(m,m')$-th element is the dot product
$\pi(x_m)^{\top} \pi(x_{m'}) \in [0, 1]$
and represents the similarity of the two distributions. 
Each element of $\mathbf{\Pi}^{\top} \mathbf{\Pi}$ takes a maximum value of 1 when the two distributions are identical one-hot distributions and a minimum value of 0 when they have no overlap.
Therefore, the Frobenius norm $\lVert\mathbf{\Pi}^{\top} \mathbf{\Pi} \rVert_F$ takes a large value when the similarity of the distributions in a mini-batch is high, whereas it takes a small value when the similarity is low.
Using this property, \citet{lin2017self} proposed to minimize 
$\lVert \mathbf{\Pi}^{\top} \mathbf{\Pi}- \mathbf{I} \rVert_F$ as a penalty term to reduce the similarity of self-attention maps.
$\mathbf{I} \in \mathbb{R}^{M \times M}$ is an identity matrix to encourage $\pi(x_m)$ to be one-hot.

We use this penalty term with the following modifications.
First, the penalty cannot be minimized to zero when the mini-batch size $M$ exceeds the number of experts $K$.
During joint minimization with the classification loss $\mathcal{L}_C$, forcing the minimization of the never-zero penalty could lead models too far away from the optimal solution for $\mathcal{L}_C$.
To avoid this, we consider that in the $m$-th row of $\mathbf{\Pi}^{\top} \mathbf{\Pi} - \mathbf{I}$, the corresponding expert for $x_m$ captures the same latent features among the top-$\ell$ elements (instances).
We then exclude those elements to allow such multi-instance assignments.
Let $d_{\ell}: \mathbb{R}^{M \times M} \rightarrow \mathbb{R}^{M \times M}$ be a function to drop out the elements with the top-$\ell$ values in each row to 0 (Algorithm~\ref{alg:top-l dropout}).
We minimize $\lVert d_{\ell}(\mathbf{\Pi}^{\top}\mathbf{\Pi} - \mathbf{I}) \rVert_F$.

Second, the penalty varies highly depending on the batch size $M$, as the Frobenius norm takes the sum, not the mean, of the squares of the matrix elements.
To search for weighting hyperparameters robust to changes in $M$, we    normalize the penalty into [0, 1].
We divide the penalty by $\lVert d_{\ell}(\mathbf{J} - \mathbf{I}) \rVert_F$, where $\mathbf{J} \in \mathbb{R}^{M \times M}$ is a matrix of ones.

\begin{algorithm}[t]
\caption{Row-wise top-$\ell$ dropout}
\label{alg:top-l dropout}
\SetAlgoLined
\KwIn{Square matrix $\mathbf{M} \in \mathbb{R}^{M \times M}$ and $\ell$}
\KwOut{Square matrix $\mathbf{M} \in \mathbb{R}^{M \times M}$}
\For{$i = 1, \cdots, M$}{
    $\mathbf{m} \leftarrow \mathbf{M}_i$ \algorithmiccomment{$\mathbf{M}_i \in \mathbb{R}^M$ is the $i$-th row vector of $\mathbf{M}$} \\
    $\mathrm{Sort}(\mathbf{m})$ \algorithmiccomment{Sort in descending order} \\
    $\mathbf{m}[:\ell] \leftarrow \mathbf{0}$ \algorithmiccomment{$\mathbf{0} \in \mathbb{R}^{\ell}$, drop out top-$\ell$} \\
    $\mathbf{M}_i \leftarrow \mathbf{m}$ \\
}
\KwRet{$\mathbf{M}$}
\end{algorithm}

Taking all of these together, we define our penalty term as follows:
\begin{align}
    \label{eq:penalty term}
    \mathcal{L}_R (\bm{\theta}) &= \frac{\lVert d_{\ell}(
    \mathbf{\Pi}^{\top}\mathbf{\Pi} - \mathbf{I}) \rVert_F}{\lVert d_{\ell}(\mathbf{J} - \mathbf{I}) \rVert_F}.
\end{align}
The final loss is defined using the weighting hyperparameters $\lambda$ as follows:
\begin{align}
    \label{eq:final loss}
    \mathcal{L} (\bm{\theta}) &= \mathcal{L}_C (\bm{\theta}) + \lambda \mathcal{L}_R (\bm{\theta}).
\end{align}

Previous studies on MoE observed that assignment was concentrated on the same few experts and proposed penalty terms to balance the assignment among experts \citep{shazeer2017outrageously,lepikhin2021gshard,fedus2022switch,fedus2022review}.
However, these penalty terms were proposed for text generation models and cannot be applied directly to the classification model in consideration.
Notably, the penalty terms encourage balanced, uniform assignments but do not encourage diverse assignments that vary among groups of instances.

\subsection{Inference Phase: Post-Hoc Control for Risk Minimization under Uncertainty}
\label{sec:control}
The problem of shortcuts emerges upon the distribution shifts where some latent features are no longer associated with labels.
In this subsection, we consider controlling the mixture weights to minimize the risk under the OOD circumstances where we do not know which latent features to rely on during inference.
For this control, we suppose that different experts capture those latent features with small overlaps, as encouraged in the training (see Section~\ref{sec:router constraint}: penalty term).
However, \emph{note that this is not a strict requirement}, and moderate differences may be sufficient.
The point is that not all experts depend on the same latent features.
We introduce two post-hoc operations on $\pi$ to ensure predictions remain robust to such shifts.
The operations replace the estimated $\pi$ with $\pi^*$ according to the theory of risk minimization under uncertainty.

\paragraph{Uniform Weighting.}
\label{sec:uniform}
The simplest way to obtain robustness to variation is to use a uniform distribution.
Assuming that all experts are equally good, a simple way to obtain robustness to the unknown shifts is to consider the expert's predictions equally.
We replace the estimated ${\pi}$ with a uniform distribution as follows:
\begin{align}
    y^{\ast} &= \argmax_{y \in \mathcal{Y}} \eqnmarkbox[eqred]{R}{K^{-1}} \sum_{k=1}^{K} \eqnmarkbox[eqgreen]{E}
    {p^{k}(y|x)}. 
\end{align}
\annotate[yshift=-0.4em]{below, right}{R}{$\pi^*$}\\
This operation is equivalent to taking the mean of $p^{k} (y|x)$ across $K$ experts.

\paragraph{Argmin Weighting.}
\label{sec:argmin}
In the worst-case scenario, the assumption that all experts are equally good does not hold.
An alternative approach to minimize the risk of erroneous predictions in this case is to determine the mixture weights by considering the expert model's predictions, as follows:
\begin{align}
    &
    y^{\ast} = \argmax_{y \in \mathcal{Y}} \sum_{k=1}^{K} 
    \eqnmarkbox[eqgreen]{E}{p^{k}(y | x)} 
    \eqnmarkbox[eqred]{R}{\mathbbm{1} \{k^{\ast}(x)=k\}}, \hspace{-0.2cm} \\
    &
    k^{\ast}(x) = \argmin_{k \in \mathcal{K}} p^{k}(y|x),
\end{align}
\annotate[yshift=-0.4em]{below, right}{R}{$\pi^*$}\\
where 
$\mathcal{K} = \{ 1, 2, ..., K \}$
and 
$\mathbbm{1}\{ \cdot \}$ is the indicator function.
This operation first selects the expert that minimizes the probability $p^{k}(y|x)$ over a set of $K$ experts for each label, and then chooses the label that maximizes the resulting probability.
See Figure~\ref{fig:argmin} in Appendix~\ref{sec:appx figures} for an example.

\paragraph{Derivation of the Operations.}
\label{sec:derivation}
The remainder of this subsection further explains the principles behind the prediction rules introduced earlier, with a focus on risk minimization.
This perspective is rooted in the classical statistical decision-making framework
\citep[see][]{Wald1950WALSDF,berger1985statistical}.

We consider a prediction function $\delta: \mathcal{X} \to \mathcal{Y}$
and define $\mathcal{D}$ as a collection of such measurable prediction functions.
To evaluate the prediction,
we employ the 0-1 loss $L: \mathcal{Y} \times \mathcal{Y} \to \{0,1\}$, which measures classification error as follows: 
\begin{align}
    L\big(y, \delta(x) \big)
    &= 
1 -  \mathbbm{1}\{  y=\delta(x) \}.
\end{align}
Since $x$ and $y$ are random variables,
the value of the loss is a random quantity. 

We consider the expected value of the loss $L\big(y, \delta(x) \big)$
and call it the risk. 
Given that the mixture weights $\pi$ are the essential elements for our analysis,
we explicitly state the dependency of the risk on the weights.
We denote the risk function $R: \mathcal{P} \times \mathcal{D} \to \mathbb{R}$, given by
\begin{align}
    R(\pi, \delta)
    &= 
    1-
    \mathbb{E}_{x} 
    \Bigg[ 
    \sum_{k=1}^{K}
    p^{k}
    \big(
      \delta(x) |x
    \big)
    \pi_{k}(x)
    \Bigg], 
\end{align}
where $\mathbb{E}_{x}[\cdot]$ represents the expectation operator over the outcomes of the random variable $x$.

If the mixture weights $\pi$ are known, then
it can be easily shown that
the risk $R(\pi, \delta)$ is minimized by  
\begin{align}
    \delta_{\pi}^{\ast}(x)
    &= \argmax_{y \in \mathcal{Y}}
    \sum_{k=1}^{K}
    p^{k}
    \big(
      y |x
    \big)
    \pi_{k}(x). 
\end{align}
That is, the optimal prediction selects an element of $\mathcal{Y}$ corresponding to the maximum conditional probability.
When the training and evaluation data share the same joint distributions of $y$ and $x$, we can apply the above classification rule with the estimated mixture weights.
However, this does not hold for OOD data.

We propose two approaches to deal with the case of unknown mixture weights. 
First, we assume that while we do not know the correct mixture weights, all experts are equally good.
Then, we can take uniform weights for the mixture weights and minimize the risk by 
\begin{align}
    \delta_{u}^{\ast}(x)
    &= \argmax_{y \in \mathcal{Y}}
    \frac{1}{K}
    \sum_{k=1}^{K}
    p^{k}
    \big(
      y|x
    \big).
\end{align}

Another approach takes a strategy of prudence, focusing on the worst-case scenario to guarantee the most favorable outcome among these least desirable possibilities. Consequently, the prediction performs uniformly well across the mixture set $\mathcal{P}$, aligning with the minimax principle.
More precisely, the minimax problem is written as   
\begin{align}
    \min_{\delta \in \mathcal{D}} \max_{\pi \in \mathcal{P}} R(\pi, \delta)
\end{align}
which is solved by the following prediction rule 
\begin{align}
    \delta_{}^{\ast}(x)
    &= 
    \argmax_{y \in \mathcal{Y}}
    \min_{k \in \mathcal{K}}
    p^{k}
    \big(
      y|x
    \big).
\end{align}
The above prediction rule is a maximin criterion where one selects the element of $\mathcal{Y}$ that maximizes the minimum conditional probability.

\section{Experiments}
\label{sec:experiments}
Our goal is to achieve predictions robust to distribution shifts related to shortcuts.
In this section, we test whether the proposed post-hoc control improves performance on those OOD tests and analyze the mechanism based on our assumption.

\subsection{Setup}
\label{sec:setup}
This subsection describes the experimental setup.
Please refer to Appendix~\ref{sec:appx setup} for further details.

\begin{table*}[t]
\centering
\begin{adjustbox}{max width=0.9\textwidth}
\begin{tabular}{llccccccccc}
\toprule
\multicolumn{1}{c}{\multirow{2}*{\emph{Fix}}} & \multicolumn{1}{c}{\multirow{2}*{\emph{Search}}} & \multicolumn{3}{c}{\textbf{MNLI}} & \multicolumn{3}{c}{\textbf{QQP}} & \multicolumn{3}{c}{\textbf{FEVER}} \\
\cmidrule(lr){3-5}
\cmidrule(lr){6-8}
\cmidrule(lr){9-11}
 &  & $\mathcal{L}_{C}$ & $\mathcal{L}_{R}$ & $\mathcal{L}_{C} + \mathcal{L}_{R}$ & $\mathcal{L}_{C}$ & $\mathcal{L}_{R}$ & $\mathcal{L}_{C} + \mathcal{L}_{R}$ & $\mathcal{L}_{C}$ & $\mathcal{L}_{R}$ & $\mathcal{L}_{C} + \mathcal{L}_{R}$ \\
\midrule
\multirow{3}*{$\mathrm{1st:\ }\lambda=0.0$}
 & $K=5$ & $0.706$ & $0.379$ & $1.084$ & $0.378$ & $0.299$ & $0.677$ & $0.440$ & $0.674$ & $1.114$ \\
 & $K=10$ & $0.433$ & $0.127$ & $\mathbf{0.560}$ & $0.415$ & $0.505$ & $0.920$ & $0.392$ & $0.417$ & $\mathbf{0.809}$ \\
 & $K=15$ & $0.501$ & $0.301$ & $0.802$ & $0.253$ & $0.245$ & $\mathbf{0.498}$ & $0.466$ & $0.367$ & $0.832$ \\
\cmidrule(lr){3-5}
\cmidrule(lr){6-8}
\cmidrule(lr){9-11}
\rowcolor{backgray}
 &  & \multicolumn{3}{c}{$\bm{K^*=10}$} & \multicolumn{3}{c}{$\bm{K^*=15}$} & \multicolumn{3}{c}{$\bm{K^*=10}$} \\
\midrule
\multirow{3}*{$\mathrm{2nd:\ }K=K^*$}
 & $\lambda=0.0$ & $0.433$ & $0.127$ & $0.560$ &  $0.253$ & $0.245$ & $0.498$ & $0.392$ & $0.417$ & $0.809$ \\
 & $\lambda=0.5$ & $0.437$ & $0.022$ & $\mathbf{0.459}$ & $0.333$ & $0.010$ & $0.343$ & $0.523$ & $0.128$ & $0.651$ \\
 & $\lambda=1.0$ & $0.666$ & $0.005$ & $0.671$ & $0.332$ & $0.003$ & $\mathbf{0.335}$ & $0.416$ & $0.132$ & $\mathbf{0.548}$ \\
\cmidrule(lr){3-5}
\cmidrule(lr){6-8}
\cmidrule(lr){9-11}
\rowcolor{backgray}
 &  & \multicolumn{3}{c}{$\bm{\lambda^*=0.5}$} & \multicolumn{3}{c}{$\bm{\lambda^*=1.0}$} & \multicolumn{3}{c}{$\bm{\lambda^*=1.0}$} \\
\bottomrule
\end{tabular}
\end{adjustbox}
\caption{
Results of the two-stage hyperparameter search for the number of experts $K$ and the loss-weighting value $\lambda$.
$K^*$ and $\lambda^*$ are the optimal values that minimize $\mathcal{L}_C + \mathcal{L}_R$ on ID dev.
All results are the average of two runs with different seeds.
}
\label{tab:val loss}
\end{table*}

\paragraph{Datasets.}
\label{sec:datasets}
In accordance with previous research on shortcut mitigation, we experimented with three NLU datasets.
These are popular datasets but are all reported to induce shortcuts, and OOD test data were later created that cannot be correctly classified by the shortcuts.
Each dataset consists of training data, validation data drawn from the same distribution as the training data (\textbf{ID dev}), and test data where the correlation between some latent features and labels changed adversarially (\textbf{OOD test}).
Following previous studies in comparison, we evaluate the accuracy.

\textbf{MNLI} \citep{williams-etal-2018-broad} is a dataset for natural language inference (NLI) across multiple genres.
Given a pair of premise and hypothesis sentences, the task is to classify the relationship between the two sentences into one of three labels: \emph{entailment}, \emph{contradiction}, or \emph{neutral}.
In MNLI, a shortcut arises from a spurious correlation between the word overlap of input sentences and target labels.
We used its matched development set as ID dev and \textbf{HANS} \citep{mccoy-etal-2019-right} as OOD test.
\textbf{QQP} is a dataset for paraphrase identification.
The task is to classify whether two sentences are paraphrases or not.
A shortcut also arises from a spurious correlation in the word overlap of input sentences.
We used its development set as ID dev and \textbf{PAWS} \citep{zhang-etal-2019-paws} as OOD test.
\textbf{FEVER} \citep{thorne-etal-2018-fever} is a dataset for fact verification.
Given two sentences of claim and evidence, the task is to classify the relation of the evidence toward the claim into either \emph{Supports}, \emph{Refutes}, or \emph{Not-enough-info}.
Some negative phrases in the claim sentences spuriously correlate with target labels, causing a shortcut that allows classification using only the claim sentences.
We used its development set as ID dev and \textbf{FEVER Symmetric v1 and v2} \citep{schuster-etal-2019-towards} as OOD tests.

\paragraph{Baseline and Principal Methods.}
\label{sec:baselines}
We used \textbf{BERT} (\texttt{bert-base-uncased}) \citep{devlin-etal-2019-bert} as the baseline and backbone for a fair comparison with previous studies.
We used the last layer in the position of \texttt{[CLS]} for $\mathbf{h}$ in Eq.~\eqref{eq:encoder h}.

To compare in \emph{a practical setting where only ID data are available for training and tuning \citep{pmlr-v202-yang23s}}, \emph{we reran principal methods in that setting} using their publicly available code.

\textbf{Conf-reg~$\spadesuit~_{\text{self-debias}}$} \citep{utama-etal-2020-towards} and \textbf{JTT} \citep{liu2021just} use heuristics that weak models are likely to exploit shortcuts.
Conf-reg~$\spadesuit~_{\text{self-debias}}$ reweights the loss according to predictions of a weak model while balancing the weights using predictions of a teacher model.
JTT up-weights the loss of training instances that a weak model misclassified.
\textbf{RISK} \citep{wu-gui-2022-less} considers shortcuts to be redundant features and applies feature reduction.
\textbf{EIIL} \citep{pmlr-v139-creager21a} first estimates the groups of training instances where some shortcuts are in common and then applies IRM (see Section~\ref{sec:previous approaches}) using the estimated groups.
\textbf{BAI} \citep{yu-etal-2022-interventional} extends EIIL to estimate multiple levels of groups and apply IRM multiple times accordingly.
\textbf{GroupDRO$_{\text{label-group}}$} \citep{sagawa2020distributionally} and \textbf{ReWeightCRT} \citep{Kang2020Decoupling} are reported to perform well on OOD data when the label distribution $p(y)$ is imbalanced in ID data but is uniform in OOD data \citep{pmlr-v202-yang23s}, while they do not aimed at addressing the shift in shortcuts.
GroupDRO$_{\text{label-group}}$ minimizes the loss on the worst-case class label given groups divided only by class labels, and ReWeightCRT reweights the loss with the relative frequency of class labels.

\begin{table*}[t]
\centering
\begin{adjustbox}{max width=0.9\textwidth}
\begin{tabular}{lccccccc}
\toprule
\multicolumn{8}{c}{\emph{Main Results: Tuning with ID Dev}} \\
\midrule
 & \multicolumn{2}{c}{\textbf{MNLI}} & \multicolumn{2}{c}{\textbf{{QQP}}} & \multicolumn{3}{c}{\textbf{FEVER}} \\
\cmidrule(lr){2-3}
\cmidrule(lr){4-5}
\cmidrule(lr){6-8}
 & \small{ID} & \small{OOD} & \small{ID} & \small{OOD} & \small{ID} & \multicolumn{2}{c}{\small{OOD}} \\
 & \textbf{Dev} & \textbf{HANS} & \textbf{Dev} & \textbf{PAWS} & \textbf{Dev} & \textbf{Symm. v1} & \textbf{Symm. v2} \\
\midrule
BERT & $\underline{84.4}_{\pm 0.2}$ & $55.2_{\pm 4.2}$ &  $\underline{91.5}_{\pm 0.1}$ & $36.7_{\pm 3.1}$ & $86.7_{\pm 0.2}$ & $58.5_{\pm 1.4}$ & $65.1_{\pm 1.5}$ \\
\rowcolor{backgray}
 \ \ + MoS & $\underline{84.4}_{\pm 0.1}$ & $59.4_{\pm 5.5}$ & $91.4_{\pm 0.1}$ & $34.9_{\pm 1.6}$ & $87.0_{\pm 0.5}$ & $58.9_{\pm 1.2}$ & $65.5_{\pm 1.0}$ \\
\rowcolor{backgray}
 \ \ \ \ $\rightarrow$ Uniform & $83.0_{\pm 1.0}$ & $63.6_{\pm 5.7}$ & $89.1_{\pm 2.4}$ & $47.0_{\pm 8.6}$ & $\underline{87.6}_{\pm 1.2}$ & $\underline{\mathbf{62.2}}_{\pm 0.7}$ & $\underline{\mathbf{68.2}}_{\pm 1.0}$ \\
\rowcolor{backgray}
 \ \ \ \ $\rightarrow$ Argmin & $81.0_{\pm 3.1}$ & $\underline{\mathbf{67.2}}_{\pm 4.6}$ & $83.8_{\pm 7.3}$ & $\underline{\mathbf{55.7}}_{\pm 8.5}$ & $85.3_{\pm 6.8}$ & $61.8_{\pm 1.2}$ & $67.4_{\pm 2.2}$ \\
\midrule
Conf-reg~$\spadesuit~_{\text{self-debias}}$ & $84.5_{\pm 0.2}$ & $63.7_{\pm 2.4}$ & $90.5_{\pm 0.2}$ & $31.0_{\pm 1.7}$ & $87.1_{\pm 0.7}$ & $59.7_{\pm 1.3}$ & $66.5_{\pm 1.1}$ \\
Conf-reg~$\spadesuit~_{\text{self-debias}}^{\text{last}}$ & $84.5_{\pm 0.2}$ & $63.7_{\pm 2.4}$ & $90.5_{\pm 0.2}$ & $31.0_{\pm 1.7}$ & $86.7_{\pm 0.4}$ & $59.3_{\pm 1.2}$ & $65.9_{\pm 1.1}$ \\
{JTT} & $80.7_{\pm 0.3}$ & $57.3_{\pm 2.2}$ & $89.4_{\pm 0.2}$ & $36.0_{\pm 0.6}$ & $82.7_{\pm 1.1}$ & $53.0_{\pm 2.6}$ & $60.3_{\pm 2.6}$ \\
{RISK} & $83.9_{\pm 0.3}$ & $56.3_{\pm 4.2}$ & $90.5_{\pm 0.1}$ & $34.8_{\pm 3.2}$ & $87.6_{\pm 0.8}$ & $58.9_{\pm 2.6}$ & $65.9_{\pm 1.6}$ \\
{EIIL} & $83.9_{\pm 0.2}$ & $61.5_{\pm 2.4}$ & $91.1_{\pm 0.2}$ & $31.0_{\pm 0.6}$ & $86.8_{\pm 1.1}$ & $56.2_{\pm 1.9}$ & $63.8_{\pm 1.7}$ \\
\ \ + {BAI} & $83.7_{\pm 0.2}$ & $62.0_{\pm 2.1}$ & $91.2_{\pm 0.2}$ & $31.2_{\pm 0.3}$ & $86.3_{\pm 1.2}$ & $56.0_{\pm 2.1}$ & $63.6_{\pm 1.9}$ \\[0.05cm]
\cdashline{1-8}
{GroupDRO$_{\text{label-group}}$} & $84.3_{\pm 0.3}$ & $57.7_{\pm 2.9}$ & $\mathbf{91.6}_{\pm 0.1}$ & $34.6_{\pm 3.7}$ & $\mathbf{89.3}_{\pm 0.2}$ & $62.1_{\pm 1.1}$ & $67.9_{\pm 1.3}$ \\
{ReWeightCRT} & $\mathbf{84.6}_{\pm 0.1}$ & $55.8_{\pm 0.3}$ & $91.5_{\pm 0.0}$ & $32.0_{\pm 0.3}$ & $88.5_{\pm 0.0}$ & $61.3_{\pm 0.4}$ & $66.9_{\pm 0.2}$ \\
\bottomrule
\end{tabular}
\end{adjustbox}
\caption{
Main results.
The results of our method are \colorbox{backgray}{colored in the background}.
All the scores are shown in the mean and standard deviation of five runs with different seeds.
The highest \emph{mean} scores are shown in \textbf{bold}, and the highest \emph{mean} scores within the baseline and our method are \underline{underlined}.
}
\label{tab:main results}
\end{table*}

\paragraph{Hyperparameters.}
\label{sec:hyperparameters}
As Eq.~\eqref{eq:final loss} shows, the optimal model for the proposed method is one that can accurately classify $x$ and output diverse $\pi(x)$.
Therefore, we define the optimal hyperparameters for the proposed method as those that minimize the sum of the two losses ($\mathcal{L}_C + \mathcal{L}_R$) on \textbf{ID dev}.

In the proposed method, the number of experts $K$ in Eq.~\eqref{eq:MoS}, the number of row-wise dropouts $\ell$ in Eq.~\eqref{eq:penalty term}, and the loss-weighting value $\lambda$ in Eq.~\eqref{eq:final loss} are model-specific hyperparameters.
We explored the values of $K \in \{5, 10, 15\}$ and $\lambda \in \{0.0, 0.5, 1.0\}$.
For an efficient search, we conducted a two-stage search.
At the first stage, we fixed $\lambda=0$ and determined $K^*$ that naturally fit the data.
Then, we searched for the optimal balance of losses $\lambda$ under $K^*$.
Table~\ref{tab:val loss} shows the results of the hyperparameter search.
Across settings, the value of $\ell$ was set to be the smallest value in $2^n$ that satisfies $\min (K) \cdot \ell \geq M$.
This ensures that $\mathcal{L}_R$ in each mini-batch of size $M$ can be zero in all settings when $\pi$ is maximally diverse: when a different expert is allocated to every $\ell$ instances with probability one.
We used parallel processing of two mini-batches of $M=32$ each and $\min (K)=5$, so we set $\ell=8$ to satisfy the condition.
Regarding epochs, we set the training epoch to 10 and the learning rate to 2e-5 for all datasets and select the best epoch on ID dev scores without applying post-hoc control.

When rerunning the comparison methods, we set all hyperparameters to the values specified in the papers or the official implementation, except for an annealing hyperparameter $\alpha$ of Conf-reg~$\spadesuit~_{\text{self-debias}}$, as it was tuned on OOD tests.
We took the best epoch on ID dev for all the methods and the best $\alpha$ on ID dev for Conf-reg~$\spadesuit~_{\text{self-debias}}$.

\subsection{Results}
\label{sec:results}
As the main results, we demonstrate that our post-hoc control over the experts achieves robust predictions on OOD test data.
Table~\ref{tab:main results} shows the results in the setting where no shortcut is pre-identified.
\textbf{BERT} is the baseline, \textbf{+ MoS} is our mixture model, and \textbf{$\bm{\rightarrow}$~Uniform~/~Argmin} performs the post-hoc control on the mixture model.
Since scores on the OOD tests have been reported to have high variance, all the results are shown in the mean and standard deviation of five runs with different seeds in accordance with previous studies.
We observe that in all datasets, our post-hoc control significantly improves performance on the OOD tests from the baseline and MoS.

The comparison methods do not improve performance on the OOD tests much when tuned solely with ID data,\footnote{Conf-reg~$\spadesuit~_{\text{self-debias}}$ reported taking the last epoch of arbitrarily determined epochs rather than ID dev best epoch.
We also reported the performance of the last-epoch models (Conf-reg~$\spadesuit~_{\text{self-debias}}^{\text{last}}$), but we found that this practice did not work well when its annealing hyperparameter $\alpha$ (see Section~\ref{sec:hyperparameters}) was tuned solely with ID data.
} which is consistent with the observation in \citet{pmlr-v202-yang23s}.
As an exception, GroupDRO$_{\text{label-group}}$ and ReWeightCRT perform well on FEVER, where the difference from our method is marginal considering the standard deviation.
This is because the label distribution of FEVER shifts as these methods suppose,\footnote{The label distribution of FEVER is approximately Supports:Refutes:Not-enough-info = 2:1:2 in the training data but Supports:Refutes:Not-enough-info = 1:1:0 in both ID dev and OOD test.
Thus, supposing the flat label distribution for Supports and Refutes improves the OOD performance even without addressing shortcuts.
} which is also consistent with the observation in \citet{pmlr-v202-yang23s}.
However, they do not improve the OOD performance on MNLI and QQP, which have no such label distribution shift.
In contrast, our method does not exploit assumptions on label distribution shifts but consistently improves the OOD performance across all the datasets.

\subsection{Analyses}
\label{sec:analyses}
Now, we turn to the mechanism behind our method's robust performance and analyze the mixture model based on our assumption.

\begin{table}[t]
\centering
\begin{adjustbox}{max width=0.9\columnwidth}
\begin{tabular}{lcc}  
\toprule
 & $\mathcal{L}_{R}$ & $\Delta_{\textrm{MoS} \rightarrow \textrm{Argmin}}$ \\
\midrule
\rowcolor{backgray}
\texttt{MNLI} &  & \\
\textbf{Train} & $0.020_{\pm 0.014}$ & - \\
\textbf{Dev} & $0.017_{\pm 0.009}$ &  $-3.4_{\ (84.4 \rightarrow 81.1)}$ \\
\textbf{HANS} & $\mathbf{0.633}_{\pm 0.075}$ & $\mathbf{+7.8}_{\ (59.4 \rightarrow 67.2)}$ \\
\midrule
\rowcolor{backgray}
\texttt{QQP} &  & \\
\textbf{Train} & $0.002_{\pm 0.001}$ & - \\
\textbf{Dev} & $0.003_{\pm 0.001}$ & $-7.6_{\ (91.4 \rightarrow 83.8)}$ \\
\textbf{PAWS} & $\mathbf{0.329}_{\pm 0.133}$ & $\mathbf{+20.8}_{\ (34.9 \rightarrow 55.7)}$ \\
\midrule
\rowcolor{myred}
\texttt{FEVER} &  & \\
\textbf{Train} & $0.010_{\pm 0.001}$ & - \\
\textbf{Dev} & $\mathbf{0.136}_{\pm 0.027}$ & $-1.7_{\ (87.0 \rightarrow 85.3)}$ \\
\textbf{Symm. v1} & $0.082_{\pm 0.016}$ & $\mathbf{+2.9}_{\ (58.9 \rightarrow 61.8)}$ \\
\textbf{Symm. v2} & $0.087_{\pm 0.028}$ & $+1.9_{\ (65.5 \rightarrow 67.4)}$ \\
\bottomrule
\end{tabular}
\end{adjustbox}
\caption{
The value of $\mathcal{L}_R$ and difference in before and after performing the post-hoc control ($\Delta_{\textrm{MoS} \rightarrow \textrm{Min}}$).
The scores were obtained with five runs of different seeds.
We \textbf{bold} the worst \emph{mean} score of $\mathcal{L}_R$ and the best gain in $\Delta_{\textrm{MoS} \rightarrow \textrm{Min}}$.
}
\label{tab:router loss}
\end{table}

\paragraph{Analysis 1: Penalty Term $\mathcal{L}_R$ in ID and OOD Data.}
\label{sec:analysis penalty}
We first analyze the penalty term $\mathcal{L}_R$, the essential statistic of our mixture model.
Recall that $\mathcal{L}_R$ encourages the router to assign different inputs to different experts, assuming different inputs have some difference in their latent features.\footnote{
While not an inevitable consequence of this objective nor a requirement for our method, we analyzed how different experts output different predictions.
Figure~\ref{fig:heatmap prediction} in Appendix~\ref{sec:appx figures} shows a significant variance between experts' predictions.}
In other words, $\mathcal{L}_R$ measures the sensitivity to the difference in inputs.
Drawing an inference from this, we expect that the value of $\mathcal{L}_R$ differs in the shifts related to latent features, that is, shifts between ID and OOD data we address.

Table~\ref{tab:router loss} shows the value of $\mathcal{L}_R$ on the ID and OOD datasets.\footnote{HANS is sorted by the type of shortcuts, so we shuffled the order before computing $\mathcal{L}_R$. We did not observe this kind of sorted pattern in the other datasets.}
In all the datasets, the values of $\mathcal{L}_R$ differ significantly between ID and OOD data, indicating that $\mathcal{L}_R$ is sensitive to the distribution shifts in these data.

From a practical perspective, this sensitivity may provide an advantage.
We can compute $\mathcal{L}_R$ during inference since its computation does not require annotated labels either in training or inference.
Therefore, during inference, we can determine which data to perform the post-hoc control on by looking at how different $\mathcal{L}_R$ is from that on ID data.
While the post-hoc control decreases the ID dev scores, MoS performs the same as the baseline on the ID dev, regardless of the training with the penalty term (Table~\ref{tab:main results}).
Thus, adaptively applying the post-hoc control enables handling both ID and OOD data.
This adaptive use is an advantage over previous methods, which only obtain a single model fitted to either OOD or ID data.
However, note that it is limited to when involving a major shift in the distribution of latent features.
Since we do not precisely know the threshold for how much difference should be regarded as a threatening shift, it may be difficult to determine in data such as FEVER, where the difference is significant but relatively small.

Interestingly, FEVER differs from the others in how $\mathcal{L}_R$ changes between ID and OOD data.
While the others have lower $\mathcal{L}_R$ on ID data and higher $\mathcal{L}_R$ on OOD data, the opposite is true on FEVER.
This suggests that the mixture model does not model latent features well on FEVER, and in fact, the performance improvement by performing the post-hoc control is relatively small on FEVER ($\bm{\Delta}_{\textbf{MoS} \bm{\rightarrow} \textbf{Argmin}}$).
Shortcuts in FEVER depend on very local patterns: particular phrases contained only in claim sentences.
Our method uses the highly abstracted final-layer features of BERT and may not be good at successfully isolating the effects of the local patterns.
The features $\mathbf{h}$ in Eq.~\eqref{eq:encoder h} can be modified arbitrarily, so we leave more effective encoding methods for future work.

\begin{figure*}[t]
\begin{minipage}{0.45\textwidth}
    \begin{subfigure}{\textwidth}
        \centering
        \includegraphics[width=\textwidth,keepaspectratio]{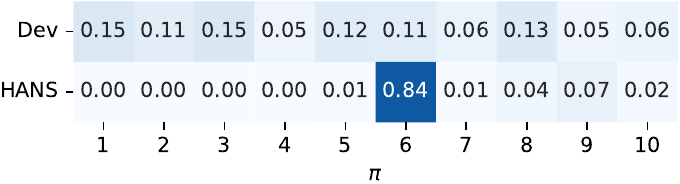}
        \caption{MNLI}
        \label{fig:heatmap mnli}
    \end{subfigure}
    \begin{subfigure}{\textwidth}
        \centering
        \includegraphics[width=\textwidth,keepaspectratio]{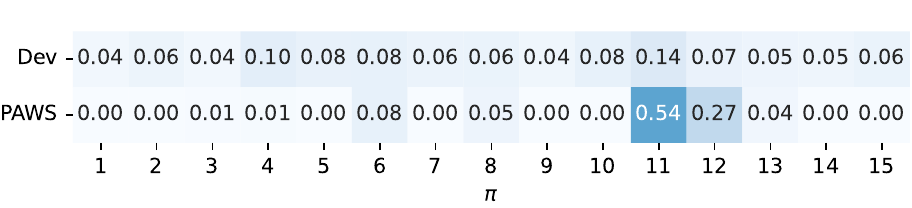}
         \caption{QQP}
         \label{fig:heatmap qqp}
    \end{subfigure}
\end{minipage}
\hfill
\begin{minipage}{0.54\textwidth}
    \begin{subfigure}{\textwidth}
        \centering
        \includegraphics[width=\textwidth,keepaspectratio]{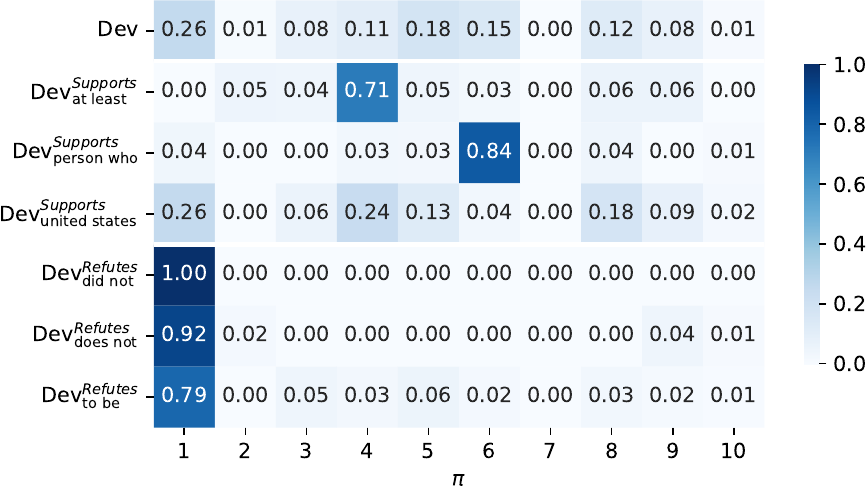}
        \caption{FEVER}
        \label{fig:heatmap fever}
    \end{subfigure}
\end{minipage}
\caption{
The mixture weights averaged on each split of the datasets.
Each split, excluding the ID dev (Dev), has its own dominant feature.
For FEVER, Dev$^{\emph{\text{label}}}_{\text{bigram}}$ consists of the instances in ID dev that contain the bigram reported to strongly correlate with the label \citep{schuster-etal-2019-towards}.
Here, no post-hoc control is performed on the mixture weights.
}
\label{fig:heatmap}
\end{figure*}

\paragraph{Analysis 2: Captured Latent Features and their Interpretability.}
\label{sec:analysis interpretability}
Our post-hoc control supposes that different experts capture different latent features to some extent.
The robust performance of our post-hoc control supports this assumption but not directly.
Toward direct validation, we analyze which experts a particular feature is assigned to.

To this end, we use some data splits in which a specific feature, known as a shortcut, is dominant.
For MNLI and QQP, high word overlap is a dominant feature in HANS and PAWS, while their creation process is different.
The sentence pairs in HANS were created by replacing or partially deleting some parts of premise sentences, while those in PAWS by word swapping, back translation, and human post-processing.
However, FEVER has no such split where a single feature is dominant.
To obtain such splits in FEVER, we extracted instances that contain frequent bigrams that are reported to strongly correlate with labels.
The bigrams ``at least'', ``person who'', and ``united states'' strongly correlate with the \emph{Supports} label and ``did not'', ``does not'', and ``to be'' with the \emph{Refutes} label \citep{schuster-etal-2019-towards}.\footnote{We omitted \emph{Not-Enough-Info} labels since the FEVER ID dev and OOD tests have no instances with this gold label.}
We created six splits from FEVER ID dev.
All the features above are known to be shortcuts.

Figure~\ref{fig:heatmap} shows the mixture weights averaged on each split of the datasets.
Note that no post-hoc control is performed on the mixture weights here.
Overall, a single or a few experts dominate the mixture weights in the splits where specific features are dominant: HANS, PAWS, and the newly created FEVER splits.
We also observe that the dominant experts differ among the FEVER splits.
Although the features under analysis are limited, this behavior of the mixture weights aligns with our assumption that different experts capture different latent features.
As an exception, there are no dominant experts in the data split of ``united states'', and the same expert is dominant in the data splits of ``did not'', ``does not'', and ``to be.''
However, note that the assumption does not need to hold completely (see Section~\ref{sec:control}), and such local bigram features may be exceptionally difficult to capture in the current encoding (see Section~\ref{sec:analysis penalty}).
Taking these points into account, we consider that the results as a whole support the assumption.

Figure~\ref{fig:heatmap} also suggests that the mixture weights provide some degree of interpretability: Instances assigned to the same expert are likely to have the same features in common.
Although our mixture model does not specify captured features by itself, the suggested interpretability may allow us to discover new prominent features in data by analyzing the commonalities of the instances assigned to the same expert.
The discovery of commonalities in each expert will serve to support the assumption further.
We leave this direction as our future work.

\begin{table}[t]
\centering
\tabcolsep 3pt
\begin{adjustbox}{max width=1\columnwidth}
\begin{tabular}{lcccc}
\toprule
 & \multicolumn{2}{c}{MoS} & \multicolumn{1}{c}{$\rightarrow$ Uniform} & \multicolumn{1}{c}{$\rightarrow$ Argmin} \\
\cmidrule(lr){2-3}
\cmidrule(lr){4-4}
\cmidrule(lr){5-5}
 & \textbf{Dev} & \textbf{HANS} & \textbf{HANS} & \textbf{HANS} \\
\midrule
\rowcolor{backgray}
$\bm{K = 10}$ & $\mathbf{84.4}_{\pm 0.1}$ & $59.4_{\pm 5.5}$ & $63.6_{\pm 5.7}$ & $\mathbf{67.2}_{\pm 4.6}$ \\
${K = 5}$ & $84.3_{\pm 0.2}$ & $58.8_{\pm 5.9}$ & $59.7_{\pm 5.9}$ & $60.9_{\pm 4.3}$ \\
${K = 15}$& $\mathbf{84.4}_{\pm 0.2}$ & $\mathbf{61.1}_{\pm 5.9}$ & $\mathbf{64.5}_{\pm 10.0}$ & $65.2_{\pm 6.4}$ \\
\midrule
\rowcolor{backgray}
$\bm{\lambda = 0.5}$ & $84.4_{\pm 0.1}$ & $59.4_{\pm 5.5}$ & $\mathbf{63.6}_{\pm 5.7}$ & $\mathbf{67.2}_{\pm 4.6}$ \\
${\lambda = 0.0}$ & $\mathbf{84.5}_{\pm 0.0}$ & $57.6_{\pm 4.8}$ & $60.0_{\pm 4.2}$ & $60.7_{\pm 4.4}$ \\
${\lambda = 1.0}$& $84.1_{\pm 0.3}$ & $\mathbf{60.3}_{\pm 4.1}$ & $57.0_{\pm 4.0}$ & $65.0_{\pm 3.9}$ \\
\midrule
\rowcolor{backgray}
$\bm{\ell = 8}$ & $84.4_{\pm 0.1}$ & $59.4_{\pm 5.5}$ & $\mathbf{63.6}_{\pm 5.7}$ & $\mathbf{67.2}_{\pm 4.6}$ \\
${\ell = 0}$ & $\mathbf{84.5}_{\pm 0.1}$ & $58.1_{\pm 5.2}$ & $57.6_{\pm 4.7}$ & $58.1_{\pm 4.6}$ \\
${\ell = 16}$& $\mathbf{84.5}_{\pm 0.1}$ & $\mathbf{59.8}_{\pm 2.7}$ & $61.0_{\pm 2.5}$ & $61.4_{\pm 3.5}$ \\
\midrule
\midrule
DeBERTa$_{\text{v3-large}}$ & $91.8_{\pm 0.0}$ & $66.3_{\pm 1.8}$ & $60.8_{\pm 10.3}$ & $74.4_{\pm 8.4}$ \\
\bottomrule
\end{tabular}
\end{adjustbox}
\caption{
Ablation study on MNLI.
The results of the best hyperparameters on the ID dev are \colorbox{backgray}{colored in the background}.
The scores were obtained with five runs of different seeds.
The highest \emph{mean} scores are shown in \textbf{bold}.
}
\label{tab:ablation}
\end{table}

\paragraph{Analysis 3: Ablation Study on Mixture Model.}
\label{sec:analysis ablation}
We analyzed the contribution to the performance with respect to the hyperparameters of our mixture model: the number of experts $K$, the number of row-wise dropouts $\ell$, and the loss-weighting value $\lambda$.
Table~\ref{tab:ablation} shows the ablation study on MNLI.
This table shows how performance changes by varying one of the hyperparameters from the values determined to be optimal on the ID dev.
There is little to no difference in performance on the ID dev for any given value, but the OOD performance with post-hoc control is best for nearly all the values determined to be best on the ID dev.
It is also worth noting that using our $\mathcal{L}_R$ and top-$\ell$ dropout consistently improves OOD performance better than without using them (when $\lambda$ or $\ell$ is zero).
These results indicate the effectiveness of the proposed training strategy and hyperparameter search.

We also tested DeBERTa$_{\text{v3-large}}$ \citep{he2023debertav} for the encoder $g_{\bm{\phi}}$.
It has around three times larger parameters than the BERT we used and performs better than BERT$_{\text{large}}$, RoBERTa$_{\text{large}}$ \citep{liu2019roberta}, XLNet$_{\text{large}}$ \citep{yang2019xlnet}, ELECTRA$_{\text{large}}$ \citep{clark2020electra}, etc., on MNLI \citep{he2023debertav}.
We conducted the same hyperparameter search for DeBERTa$_{\text{v3-large}}$ and found the best hyperparameters were exactly the same as BERT's.
The results show that the larger model significantly improves not only ID performance but also OOD performance.
However, there is still a gap between ID and OOD performance, and applying the proposed method further improves OOD performance.
These results indicate that even for large models, our method is effective in improving OOD performance.

\paragraph{Analysis 4: Identifiability of Finite Mixture.}
\label{sec:analysis identifiability}
The empirical results clearly demonstrate the effectiveness of our approach.
Nevertheless, another limitation of this work is that we do not provide a theoretical guarantee for our mixture model to capture latent features within the data.
This issue has previously been studied in the statistical literature and is referred to as the identification problem of finite mixtures.
See \citet{huang2012mixture,compiani2016using,xiang2019overview} for the recent development of finite mixture models.
As explained by \citet{compiani2016using} among others, the identification of a finite mixture model is accomplished when predictors have a distinct influence on both the outcome prediction and mixture weights.
Consistent with this, our penalty term $\mathcal{L}_R$ is designed to ensure the experts and router play distinct roles in determining the conditional outcome probabilities and the mixture weights.
This approach allows our model to effectively capture and reflect the significant variations found within the data.
From our empirical Analyses 1 and 3, the penalty term $\mathcal{L}_R$ is indeed understood as an important source of identifying mixture weights.
Since our main focus is the excellent performance of our approach in NLU applications, we plan to leave the theoretical analysis of identification for future work.

\section{Related Work}
\label{sec:related work}
As seen in Section~\ref{sec:shortcuts}, datasets for NLU tasks are known to have multiple shortcuts due to the simple heuristics, preferences, etc., possessed by annotators \citep{gururangan-etal-2018-annotation,geva-etal-2019-modeling}, or more fundamentally, the compositional nature of natural language \citep{gardner-etal-2021-competency}.
A number of studies have addressed the problem of shortcuts in NLU, but their primary difference lies in prior knowledge of shortcuts.

\paragraph{Known Shortcut Setting.}
\label{sec:known}
This setting allows models to know the existence and details of shortcuts in advance.
Previous studies used this prior knowledge to mitigate the identified shortcuts.

Reweighting is the basic strategy of previous methods.
They used shortcut-dependent models that only take shortcut features as input, \textit{e.g.}, word overlap~\citep{clark-etal-2019-dont,he-etal-2019-unlearn,mahabadi-etal-2020-end}.
These shortcut-dependent models let main models know which training instances cannot be predicted correctly via shortcuts and thus should be up-weighted.
\citet{xiong2021uncertainty} showed that the performance of these methods was further enhanced by calibrating the uncertainty of the shortcut-dependent models.
\citet{utama-etal-2020-mind} additionally employed a teacher model to adjust the weights so that a main model would not deviate too much from the distribution of training data.
\citet{belinkov-etal-2019-dont} and \citet{stacey-etal-2020-avoiding} trained a main model adversarially to a shortcut-dependent classifier.

\citet{izmailov2022on,kirichenko2023last} first trained a model on ID data and then re-trained its last classification layer on a small amount of OOD data, showing that this small parameter update for the ID-fitted model is enough to improve OOD performance.

Several approaches used the counterfactual framework of causal inference.
To make counterfactual predictions unaffected by shortcuts, \citet{tian2022debiasing,Niu_2021_CVPR} combined predictions of a main model and a shortcut-dependent model.
\citet{wang-culotta-2020-identifying} classified features into genuine or spurious and selected genuine features for predictions.
Others utilized identified spurious features to train model predictions to be invariant to interventions on the spurious features \citep{veitch2021counterfactual,pmlr-v151-makar22a,puli2022outofdistribution}.

The above methods effectively mitigate shortcuts but require the significant cost of careful analysis to achieve the prior knowledge of shortcuts.

\paragraph{Unknown Shortcut Setting.}
\label{sec:unknown}
The existence and details of shortcuts are generally unknown.
Another line of studies has sought a way to mitigate shortcuts without the cost of manual identification.

The basic strategy is reweighting, just as in the known shortcut setting.
To estimate the weights, previous methods have utilized the heuristics that weak models are likely to exploit shortcuts.
The weak models include models with limited capacity \citep{clark-etal-2020-learning,sanh2021learning}, models trained with a limited number of data \citep{utama-etal-2020-towards}, a single epoch \citep{du-etal-2021-towards}, or shallow layers \citep{ghaddar-etal-2021-end,wang-etal-2022-robust}.
While these methods used continuous weights, \citet{yaghoobzadeh-etal-2021-increasing} used binary weights that take the value of 1 only for training instances that a weak model misclassified.
Other approaches applied reweighting when learning to prune fully trained models \citep{meissner-etal-2022-debiasing,du-etal-2023-robustness,liu2022win}.

Some studies addressed shortcuts in the feature space by removing redundancy \citep{wu-gui-2022-less} or correlations~\citep{dou-etal-2022-decorrelate,gao-etal-2022-kernel} in the space.
These studies reported the best scores at different epochs for each of the ID validation data and OOD test data, so their results are not directly comparable to the other studies.

The above methods still require OOD test data related to pre-identified shortcuts to tune hyperparameters, as described in Section~\ref{sec:problems}.
Our method is different from them in 
that the training and tuning can be conducted solely on ID data.
In Section~\ref{sec:results}, we demonstrated that in the setting of \emph{fully unknown shortcuts} where only ID data are available, our method improves the performance on OOD data significantly better than the previous methods.

\paragraph{Additional Data.}
Other studies make use of additional data to mitigate shortcuts.
Counterfactual data augmentation is one such study.
Counterfactual data were generated using manual annotation \citep{kaushik2021explaining}, known shortcuts \citep{wu-etal-2022-generating}, or large language models \citep{wen-etal-2022-autocad,chen-etal-2023-disco}.
Other studies used human explanation \citep{stacey2022supervising,stacey-etal-2022-logical} or human gaze signals \citep{ren-xiong-2023-huaslim} as additional supervision to guide models during training.
Although effective, collecting these external data is cost-intensive and requires additional training.

\paragraph{Literature outside of NLU.}
Outside of NLU, shortcuts have been addressed in the ML literature as one of the broader OOD problems \citep{pmlr-v139-krueger21a,pmlr-v202-yang23s}.
Still, many methods used in ML and NLU tasks have the same concepts in common, such as reweighting \citep{nam2020learning,liu2021just,clark-etal-2020-learning,utama-etal-2020-towards}, IRM \citep{pmlr-v139-creager21a,yu-etal-2022-interventional}, counterfactual invariance \citep{veitch2021counterfactual,pmlr-v151-makar22a,puli2022outofdistribution}, and data augmentation \citep{pmlr-v162-yao22b,puli2022nuisances,wu-etal-2022-generating}.
As described in Section~\ref{sec:previous approaches}, IRM \citep{arjovsky2019invariant} and GroupDRO \citep{sagawa2020distributionally} are the two principal approaches.
These approaches considered the known shortcut setting, and similar to NLU literature, their follow-up approaches have sought to address shortcuts in the unknown shortcut setting \citep{nam2020learning,liu2021just,pmlr-v139-creager21a,pmlr-v162-yao22b,puli2022nuisances,izmailov2022on,kirichenko2023last}.
However, also similar to NLU literature, those follow-up approaches still require the shortcuts that shift in test data to be pre-identified in validation data \citep{pmlr-v202-yang23s}.

\section{Conclusion}
\label{sec:conclusion}
This study proposed a conceptually novel approach to address the shortcuts problem by pessimistically aggregating the mixture model's predictions at inference time.
We introduced the MoE-based model, penalty term to encourage different experts to capture different latent features, and post-hoc control for the mixture weights that is theoretically grounded in risk minimization.
The experimental results show that our method not only significantly enhances the model's robustness to shifts in shortcuts but also provides additional benefits to address the previous methods' problems: the performance trade-off between ID and OOD data and the need for OOD test or validation data to tune hyperparameters.

Our analyses provided results supporting the assumption: Different experts capture different latent features to some extent.
However, we also noted the limitations in the encoding method (Analysis 1), the tested features and interpretability (Analysis 2), and the theoretical guarantee of identifiability (Analysis 4).
Future work includes improving the encoding method to capture latent features more accurately, analyzing the instances assigned to the same expert to interpret what it captures and further support the assumption, and theoretically accounting for how the penalty term enhances identifiability.
While the focus of this study is on shortcuts, another future direction is extending our method to address a broader range of OOD problems (see Section~\ref{sec:shortcuts}).
We believe these are interesting future research departing from this study.

\section*{Acknowledgments}
\label{sec:acknowledgment}
We thank Jacob Eisenstein, who served as our TACL action editor, and the anonymous reviewers for their insightful comments.

\bibliography{anthology, tacl2023}

\begin{thebibliography}{78}
\expandafter\ifx\csname natexlab\endcsname\relax\def\natexlab#1{#1}\fi

\bibitem[{Arjovsky et~al.(2019)Arjovsky, Bottou, Gulrajani, and Lopez-Paz}]{arjovsky2019invariant}
Martin Arjovsky, L{\'e}on Bottou, Ishaan Gulrajani, and David Lopez-Paz. 2019.
\newblock \href {https://doi.org/10.48550/arXiv.1907.02893} {Invariant risk minimization}.
\newblock \emph{arXiv preprint arXiv:1907.02893v3}.

\bibitem[{Belinkov et~al.(2019)Belinkov, Poliak, Shieber, Van~Durme, and Rush}]{belinkov-etal-2019-dont}
Yonatan Belinkov, Adam Poliak, Stuart Shieber, Benjamin Van~Durme, and Alexander Rush. 2019.
\newblock \href {https://doi.org/10.18653/v1/P19-1084} {Don{'}t take the premise for granted: Mitigating artifacts in natural language inference}.
\newblock In \emph{Proceedings of the 57th Annual Meeting of the Association for Computational Linguistics}, pages 877--891, Florence, Italy. Association for Computational Linguistics.

\bibitem[{Berger(1985)}]{berger1985statistical}
James~O. Berger. 1985.
\newblock \href {https://doi.org/https://doi.org/10.1007/978-1-4757-4286-2} {\emph{Statistical Decision Theory and Bayesian Analysis}}.
\newblock Springer-Verlag, New York.

\bibitem[{Chen et~al.(2023)Chen, Gao, Bosselut, Sabharwal, and Richardson}]{chen-etal-2023-disco}
Zeming Chen, Qiyue Gao, Antoine Bosselut, Ashish Sabharwal, and Kyle Richardson. 2023.
\newblock \href {https://aclanthology.org/2023.acl-long.302} {{DISCO}: Distilling counterfactuals with large language models}.
\newblock In \emph{Proceedings of the 61st Annual Meeting of the Association for Computational Linguistics (Volume 1: Long Papers)}, pages 5514--5528, Toronto, Canada. Association for Computational Linguistics.

\bibitem[{Clark et~al.(2019)Clark, Yatskar, and Zettlemoyer}]{clark-etal-2019-dont}
Christopher Clark, Mark Yatskar, and Luke Zettlemoyer. 2019.
\newblock \href {https://doi.org/10.18653/v1/D19-1418} {Don{'}t take the easy way out: Ensemble based methods for avoiding known dataset biases}.
\newblock In \emph{Proceedings of the 2019 Conference on Empirical Methods in Natural Language Processing and the 9th International Joint Conference on Natural Language Processing (EMNLP-IJCNLP)}, pages 4069--4082, Hong Kong, China. Association for Computational Linguistics.

\bibitem[{Clark et~al.(2020{\natexlab{a}})Clark, Yatskar, and Zettlemoyer}]{clark-etal-2020-learning}
Christopher Clark, Mark Yatskar, and Luke Zettlemoyer. 2020{\natexlab{a}}.
\newblock \href {https://doi.org/10.18653/v1/2020.findings-emnlp.272} {Learning to model and ignore dataset bias with mixed capacity ensembles}.
\newblock In \emph{Findings of the Association for Computational Linguistics: EMNLP 2020}, pages 3031--3045, Online. Association for Computational Linguistics.

\bibitem[{Clark et~al.(2020{\natexlab{b}})Clark, Luong, Le, and Manning}]{clark2020electra}
Kevin Clark, Minh-Thang Luong, Quoc~V. Le, and Christopher~D. Manning. 2020{\natexlab{b}}.
\newblock \href {https://openreview.net/forum?id=r1xMH1BtvB} {Electra: Pre-training text encoders as discriminators rather than generators}.
\newblock In \emph{International Conference on Learning Representations}.

\bibitem[{Compiani and Kitamura(2016)}]{compiani2016using}
Giovanni Compiani and Yuichi Kitamura. 2016.
\newblock \href {https://doi.org/10.1111/ectj.12068} {{Using Mixtures in Econometric Models: A Brief Review and Some New Results}}.
\newblock \emph{The Econometrics Journal}, 19(3):C95--C127.

\bibitem[{Creager et~al.(2021)Creager, Jacobsen, and Zemel}]{pmlr-v139-creager21a}
Elliot Creager, Joern-Henrik Jacobsen, and Richard Zemel. 2021.
\newblock \href {https://proceedings.mlr.press/v139/creager21a.html} {Environment inference for invariant learning}.
\newblock In \emph{Proceedings of the 38th International Conference on Machine Learning}, volume 139 of \emph{Proceedings of Machine Learning Research}, pages 2189--2200. PMLR.

\bibitem[{D'Amour et~al.(2022)D'Amour, Heller, Moldovan, Adlam, Alipanahi, Beutel, Chen, Deaton, Eisenstein, Hoffman, Hormozdiari, Houlsby, Hou, Jerfel, Karthikesalingam, Lucic, Ma, McLean, Mincu, Mitani, Montanari, Nado, Natarajan, Nielson, Osborne, Raman, Ramasamy, Sayres, Schrouff, Seneviratne, Sequeira, Suresh, Veitch, Vladymyrov, Wang, Webster, Yadlowsky, Yun, Zhai, and Sculley}]{damour2022under}
Alexander D'Amour, Katherine Heller, Dan Moldovan, Ben Adlam, Babak Alipanahi, Alex Beutel, Christina Chen, Jonathan Deaton, Jacob Eisenstein, Matthew~D. Hoffman, Farhad Hormozdiari, Neil Houlsby, Shaobo Hou, Ghassen Jerfel, Alan Karthikesalingam, Mario Lucic, Yian Ma, Cory McLean, Diana Mincu, Akinori Mitani, Andrea Montanari, Zachary Nado, Vivek Natarajan, Christopher Nielson, Thomas~F. Osborne, Rajiv Raman, Kim Ramasamy, Rory Sayres, Jessica Schrouff, Martin Seneviratne, Shannon Sequeira, Harini Suresh, Victor Veitch, Max Vladymyrov, Xuezhi Wang, Kellie Webster, Steve Yadlowsky, Taedong Yun, Xiaohua Zhai, and D.~Sculley. 2022.
\newblock \href {http://jmlr.org/papers/v23/20-1335.html} {Underspecification presents challenges for credibility in modern machine learning}.
\newblock \emph{Journal of Machine Learning Research}, 23(226):1--61.

\bibitem[{Devlin et~al.(2019)Devlin, Chang, Lee, and Toutanova}]{devlin-etal-2019-bert}
Jacob Devlin, Ming-Wei Chang, Kenton Lee, and Kristina Toutanova. 2019.
\newblock \href {https://doi.org/10.18653/v1/N19-1423} {{BERT}: Pre-training of deep bidirectional transformers for language understanding}.
\newblock In \emph{Proceedings of the 2019 Conference of the North {A}merican Chapter of the Association for Computational Linguistics: Human Language Technologies, Volume 1 (Long and Short Papers)}, pages 4171--4186, Minneapolis, Minnesota. Association for Computational Linguistics.

\bibitem[{Dou et~al.(2022)Dou, Zheng, Wu, Gao, Shan, Zhang, Wu, and Huang}]{dou-etal-2022-decorrelate}
Shihan Dou, Rui Zheng, Ting Wu, SongYang Gao, Junjie Shan, Qi~Zhang, Yueming Wu, and Xuanjing Huang. 2022.
\newblock \href {https://aclanthology.org/2022.coling-1.199} {Decorrelate irrelevant, purify relevant: Overcome textual spurious correlations from a feature perspective}.
\newblock In \emph{Proceedings of the 29th International Conference on Computational Linguistics}, pages 2278--2287, Gyeongju, Republic of Korea. International Committee on Computational Linguistics.

\bibitem[{Du et~al.(2021)Du, Manjunatha, Jain, Deshpande, Dernoncourt, Gu, Sun, and Hu}]{du-etal-2021-towards}
Mengnan Du, Varun Manjunatha, Rajiv Jain, Ruchi Deshpande, Franck Dernoncourt, Jiuxiang Gu, Tong Sun, and Xia Hu. 2021.
\newblock \href {https://doi.org/10.18653/v1/2021.naacl-main.71} {Towards interpreting and mitigating shortcut learning behavior of {NLU} models}.
\newblock In \emph{Proceedings of the 2021 Conference of the North American Chapter of the Association for Computational Linguistics: Human Language Technologies}, pages 915--929, Online. Association for Computational Linguistics.

\bibitem[{Du et~al.(2023)Du, Mukherjee, Cheng, Shokouhi, Hu, and Awadallah}]{du-etal-2023-robustness}
Mengnan Du, Subhabrata Mukherjee, Yu~Cheng, Milad Shokouhi, Xia Hu, and Ahmed~Hassan Awadallah. 2023.
\newblock \href {https://aclanthology.org/2023.eacl-main.129} {Robustness challenges in model distillation and pruning for natural language understanding}.
\newblock In \emph{Proceedings of the 17th Conference of the European Chapter of the Association for Computational Linguistics}, pages 1766--1778, Dubrovnik, Croatia. Association for Computational Linguistics.

\bibitem[{Eisenstein(2022)}]{eisenstein-2022-informativeness}
Jacob Eisenstein. 2022.
\newblock \href {https://doi.org/10.18653/v1/2022.naacl-main.321} {Informativeness and invariance: Two perspectives on spurious correlations in natural language}.
\newblock In \emph{Proceedings of the 2022 Conference of the North American Chapter of the Association for Computational Linguistics: Human Language Technologies}, pages 4326--4331, Seattle, United States. Association for Computational Linguistics.

\bibitem[{Feder et~al.(2022)Feder, Keith, Manzoor, Pryzant, Sridhar, Wood-Doughty, Eisenstein, Grimmer, Reichart, Roberts, Stewart, Veitch, and Yang}]{feder-etal-2022-causal}
Amir Feder, Katherine~A. Keith, Emaad Manzoor, Reid Pryzant, Dhanya Sridhar, Zach Wood-Doughty, Jacob Eisenstein, Justin Grimmer, Roi Reichart, Margaret~E. Roberts, Brandon~M. Stewart, Victor Veitch, and Diyi Yang. 2022.
\newblock \href {https://doi.org/10.1162/tacl_a_00511} {Causal inference in natural language processing: Estimation, prediction, interpretation and beyond}.
\newblock \emph{Transactions of the Association for Computational Linguistics}, 10:1138--1158.

\bibitem[{Fedus et~al.(2022{\natexlab{a}})Fedus, Dean, and Zoph}]{fedus2022review}
William Fedus, Jeff Dean, and Barret Zoph. 2022{\natexlab{a}}.
\newblock \href {https://doi.org/10.48550/arXiv.2209.01667} {A review of sparse expert models in deep learning}.
\newblock \emph{\\arXiv preprint arXiv:2209.01667v1}.

\bibitem[{Fedus et~al.(2022{\natexlab{b}})Fedus, Zoph, and Shazeer}]{fedus2022switch}
William Fedus, Barret Zoph, and Noam Shazeer. 2022{\natexlab{b}}.
\newblock \href {http://jmlr.org/papers/v23/21-0998.html} {Switch transformers: Scaling to trillion parameter models with simple and efficient sparsity}.
\newblock \emph{Journal of Machine Learning Research}, 23(120):1--39.

\bibitem[{Gao et~al.(2022)Gao, Dou, Zhang, and Huang}]{gao-etal-2022-kernel}
SongYang Gao, Shihan Dou, Qi~Zhang, and Xuanjing Huang. 2022.
\newblock \href {https://aclanthology.org/2022.emnlp-main.275} {Kernel-whitening: Overcome dataset bias with isotropic sentence embedding}.
\newblock In \emph{Proceedings of the 2022 Conference on Empirical Methods in Natural Language Processing}, pages 4112--4122, Abu Dhabi, United Arab Emirates. Association for Computational Linguistics.

\bibitem[{Gardner et~al.(2021)Gardner, Merrill, Dodge, Peters, Ross, Singh, and Smith}]{gardner-etal-2021-competency}
Matt Gardner, William Merrill, Jesse Dodge, Matthew Peters, Alexis Ross, Sameer Singh, and Noah~A. Smith. 2021.
\newblock \href {https://doi.org/10.18653/v1/2021.emnlp-main.135} {Competency problems: On finding and removing artifacts in language data}.
\newblock In \emph{Proceedings of the 2021 Conference on Empirical Methods in Natural Language Processing}, pages 1801--1813, Online and Punta Cana, Dominican Republic. Association for Computational Linguistics.

\bibitem[{Geva et~al.(2019)Geva, Goldberg, and Berant}]{geva-etal-2019-modeling}
Mor Geva, Yoav Goldberg, and Jonathan Berant. 2019.
\newblock \href {https://doi.org/10.18653/v1/D19-1107} {Are we modeling the task or the annotator? an investigation of annotator bias in natural language understanding datasets}.
\newblock In \emph{Proceedings of the 2019 Conference on Empirical Methods in Natural Language Processing and the 9th International Joint Conference on Natural Language Processing (EMNLP-IJCNLP)}, pages 1161--1166, Hong Kong, China. Association for Computational Linguistics.

\bibitem[{Ghaddar et~al.(2021)Ghaddar, Langlais, Rezagholizadeh, and Rashid}]{ghaddar-etal-2021-end}
Abbas Ghaddar, Phillippe Langlais, Mehdi Rezagholizadeh, and Ahmad Rashid. 2021.
\newblock \href {https://doi.org/10.18653/v1/2021.findings-acl.168} {End-to-end self-debiasing framework for robust {NLU} training}.
\newblock In \emph{Findings of the Association for Computational Linguistics: ACL-IJCNLP 2021}, pages 1923--1929, Online. Association for Computational Linguistics.

\bibitem[{Gururangan et~al.(2018)Gururangan, Swayamdipta, Levy, Schwartz, Bowman, and Smith}]{gururangan-etal-2018-annotation}
Suchin Gururangan, Swabha Swayamdipta, Omer Levy, Roy Schwartz, Samuel Bowman, and Noah~A. Smith. 2018.
\newblock \href {https://doi.org/10.18653/v1/N18-2017} {Annotation artifacts in natural language inference data}.
\newblock In \emph{Proceedings of the 2018 Conference of the North {A}merican Chapter of the Association for Computational Linguistics: Human Language Technologies, Volume 2 (Short Papers)}, pages 107--112, New Orleans, Louisiana. Association for Computational Linguistics.

\bibitem[{He et~al.(2019)He, Zha, and Wang}]{he-etal-2019-unlearn}
He~He, Sheng Zha, and Haohan Wang. 2019.
\newblock \href {https://doi.org/10.18653/v1/D19-6115} {Unlearn dataset bias in natural language inference by fitting the residual}.
\newblock In \emph{Proceedings of the 2nd Workshop on Deep Learning Approaches for Low-Resource NLP (DeepLo 2019)}, pages 132--142, Hong Kong, China. Association for Computational Linguistics.

\bibitem[{He et~al.(2023)He, Gao, and Chen}]{he2023debertav}
Pengcheng He, Jianfeng Gao, and Weizhu Chen. 2023.
\newblock \href {https://openreview.net/forum?id=sE7-XhLxHA} {De{BERT}av3: Improving de{BERT}a using {ELECTRA}-style pre-training with gradient-disentangled embedding sharing}.
\newblock In \emph{The Eleventh International Conference on Learning Representations}.

\bibitem[{Huang and Yao(2012)}]{huang2012mixture}
Mian Huang and Weixin Yao. 2012.
\newblock \href {http://www.jstor.org/stable/23239605} {Mixture of regression models with varying mixing proportions: A semiparametric approach}.
\newblock \emph{Journal of the American Statistical Association}, 107(498):711--724.

\bibitem[{Izmailov et~al.(2022)Izmailov, Kirichenko, Gruver, and Wilson}]{izmailov2022on}
Pavel Izmailov, Polina Kirichenko, Nate Gruver, and Andrew~G Wilson. 2022.
\newblock \href {https://proceedings.neurips.cc/paper_files/paper/2022/file/fb64a552feda3d981dbe43527a80a07e-Paper-Conference.pdf} {On feature learning in the presence of spurious correlations}.
\newblock In \emph{Advances in Neural Information Processing Systems}, volume~35, pages 38516--38532. Curran Associates, Inc.

\bibitem[{Jacobs et~al.(1991)Jacobs, Jordan, Nowlan, and Hinton}]{jacobs1991adaptive}
Robert~A Jacobs, Michael~I Jordan, Steven~J Nowlan, and Geoffrey~E Hinton. 1991.
\newblock \href {https://doi.org/10.1162/neco.1991.3.1.79} {Adaptive mixtures of local experts}.
\newblock \emph{Neural computation}, 3(1):79--87.

\bibitem[{Kang et~al.(2020)Kang, Xie, Rohrbach, Yan, Gordo, Feng, and Kalantidis}]{Kang2020Decoupling}
Bingyi Kang, Saining Xie, Marcus Rohrbach, Zhicheng Yan, Albert Gordo, Jiashi Feng, and Yannis Kalantidis. 2020.
\newblock \href {https://openreview.net/forum?id=r1gRTCVFvB} {Decoupling representation and classifier for long-tailed recognition}.
\newblock In \emph{International Conference on Learning Representations}.

\bibitem[{Kaushik et~al.(2021)Kaushik, Setlur, Hovy, and Lipton}]{kaushik2021explaining}
Divyansh Kaushik, Amrith Setlur, Eduard~H Hovy, and Zachary~Chase Lipton. 2021.
\newblock \href {https://openreview.net/forum?id=HHiiQKWsOcV} {Explaining the efficacy of counterfactually augmented data}.
\newblock In \emph{International Conference on Learning Representations}.

\bibitem[{Kirichenko et~al.(2023)Kirichenko, Izmailov, and Wilson}]{kirichenko2023last}
Polina Kirichenko, Pavel Izmailov, and Andrew~Gordon Wilson. 2023.
\newblock \href {https://openreview.net/forum?id=Zb6c8A-Fghk} {Last layer re-training is sufficient for robustness to spurious correlations}.
\newblock In \emph{The Eleventh International Conference on Learning Representations}.

\bibitem[{Krueger et~al.(2021)Krueger, Caballero, Jacobsen, Zhang, Binas, Zhang, Priol, and Courville}]{pmlr-v139-krueger21a}
David Krueger, Ethan Caballero, Joern-Henrik Jacobsen, Amy Zhang, Jonathan Binas, Dinghuai Zhang, Remi~Le Priol, and Aaron Courville. 2021.
\newblock \href {https://proceedings.mlr.press/v139/krueger21a.html} {Out-of-distribution generalization via risk extrapolation (rex)}.
\newblock In \emph{Proceedings of the 38th International Conference on Machine Learning}, volume 139 of \emph{Proceedings of Machine Learning Research}, pages 5815--5826. PMLR.

\bibitem[{Lepikhin et~al.(2021)Lepikhin, Lee, Xu, Chen, Firat, Huang, Krikun, Shazeer, and Chen}]{lepikhin2021gshard}
Dmitry Lepikhin, HyoukJoong Lee, Yuanzhong Xu, Dehao Chen, Orhan Firat, Yanping Huang, Maxim Krikun, Noam Shazeer, and Zhifeng Chen. 2021.
\newblock \href {https://openreview.net/forum?id=qrwe7XHTmYb} {{GS}hard: Scaling giant models with conditional computation and automatic sharding}.
\newblock In \emph{International Conference on Learning Representations}.

\bibitem[{Lin et~al.(2017)Lin, Feng, dos Santos, Yu, Xiang, Zhou, and Bengio}]{lin2017self}
Zhouhan Lin, Minwei Feng, Cicero~Nogueira dos Santos, Mo~Yu, Bing Xiang, Bowen Zhou, and Yoshua Bengio. 2017.
\newblock \href {https://openreview.net/forum?id=BJC_jUqxe} {A structured self-attentive sentence embedding}.
\newblock In \emph{International Conference on Learning Representations}.

\bibitem[{Liu et~al.(2021)Liu, Haghgoo, Chen, Raghunathan, Koh, Sagawa, Liang, and Finn}]{liu2021just}
Evan~Z Liu, Behzad Haghgoo, Annie~S Chen, Aditi Raghunathan, Pang~Wei Koh, Shiori Sagawa, Percy Liang, and Chelsea Finn. 2021.
\newblock \href {https://proceedings.mlr.press/v139/liu21f.html} {Just train twice: Improving group robustness without training group information}.
\newblock In \emph{Proceedings of the 38th International Conference on Machine Learning}, volume 139 of \emph{Proceedings of Machine Learning Research}, pages 6781--6792. PMLR.

\bibitem[{Liu et~al.(2019)Liu, Ott, Goyal, Du, Joshi, Chen, Levy, Lewis, Zettlemoyer, and Stoyanov}]{liu2019roberta}
Yinhan Liu, Myle Ott, Naman Goyal, Jingfei Du, Mandar Joshi, Danqi Chen, Omer Levy, Mike Lewis, Luke Zettlemoyer, and Veselin Stoyanov. 2019.
\newblock \href {https://doi.org/10.48550/arXiv.1907.11692} {Roberta: A robustly optimized bert pretraining approach}.
\newblock \emph{arXiv preprint arXiv:1907.11692v1}.

\bibitem[{Liu et~al.(2022)Liu, Meng, Lin, Li, Fu, Cao, Wang, and Zhou}]{liu2022win}
Yuanxin Liu, Fandong Meng, Zheng Lin, Jiangnan Li, Peng Fu, Yanan Cao, Weiping Wang, and Jie Zhou. 2022.
\newblock \href {https://proceedings.neurips.cc/paper_files/paper/2022/file/7a27143ea615262a0c122eb179c9b7a6-Paper-Conference.pdf} {A win-win deal: Towards sparse and robust pre-trained language models}.
\newblock In \emph{Advances in Neural Information Processing Systems}, volume~35, pages 19189--19202. Curran Associates, Inc.

\bibitem[{Mahabadi et~al.(2020)Mahabadi, Belinkov, and Henderson}]{mahabadi-etal-2020-end}
Rabeeh~Karimi Mahabadi, Yonatan Belinkov, and James Henderson. 2020.
\newblock \href {https://doi.org/10.18653/v1/2020.acl-main.769} {End-to-end bias mitigation by modelling biases in corpora}.
\newblock In \emph{Proceedings of the 58th Annual Meeting of the Association for Computational Linguistics}, pages 8706--8716, Online. Association for Computational Linguistics.

\bibitem[{Makar et~al.(2022)Makar, Packer, Moldovan, Blalock, Halpern, and D'Amour}]{pmlr-v151-makar22a}
Maggie Makar, Ben Packer, Dan Moldovan, Davis Blalock, Yoni Halpern, and Alexander D'Amour. 2022.
\newblock \href {https://proceedings.mlr.press/v151/makar22a.html} {Causally motivated shortcut removal using auxiliary labels}.
\newblock In \emph{Proceedings of The 25th International Conference on Artificial Intelligence and Statistics}, volume 151 of \emph{Proceedings of Machine Learning Research}, pages 739--766. PMLR.

\bibitem[{McCoy et~al.(2019)McCoy, Pavlick, and Linzen}]{mccoy-etal-2019-right}
Tom McCoy, Ellie Pavlick, and Tal Linzen. 2019.
\newblock \href {https://doi.org/10.18653/v1/P19-1334} {Right for the wrong reasons: Diagnosing syntactic heuristics in natural language inference}.
\newblock In \emph{Proceedings of the 57th Annual Meeting of the Association for Computational Linguistics}, pages 3428--3448, Florence, Italy. Association for Computational Linguistics.

\bibitem[{Meissner et~al.(2022)Meissner, Sugawara, and Aizawa}]{meissner-etal-2022-debiasing}
Johannes~Mario Meissner, Saku Sugawara, and Akiko Aizawa. 2022.
\newblock \href {https://aclanthology.org/2022.emnlp-main.517} {Debiasing masks: A new framework for shortcut mitigation in {NLU}}.
\newblock In \emph{Proceedings of the 2022 Conference on Empirical Methods in Natural Language Processing}, pages 7607--7613, Abu Dhabi, United Arab Emirates. Association for Computational Linguistics.

\bibitem[{Nam et~al.(2020)Nam, Cha, Ahn, Lee, and Shin}]{nam2020learning}
Junhyun Nam, Hyuntak Cha, Sungsoo Ahn, Jaeho Lee, and Jinwoo Shin. 2020.
\newblock \href {https://proceedings.neurips.cc/paper_files/paper/2020/file/eddc3427c5d77843c2253f1e799fe933-Paper.pdf} {Learning from failure: De-biasing classifier from biased classifier}.
\newblock In \emph{Advances in Neural Information Processing Systems}, volume~33, pages 20673--20684. Curran Associates, Inc.

\bibitem[{Nguyen et~al.(2020)Nguyen, Nguyen, Chamroukhi, and McLachlan}]{nguyen2020approximation}
T.~Tin Nguyen, Hien~D. Nguyen, Faicel Chamroukhi, and Geoffrey~J McLachlan. 2020.
\newblock \href {https://doi.org/10.1080/25742558.2020.1750861} {Approximation by finite mixtures of continuous density functions that vanish at infinity}.
\newblock \emph{Cogent Mathematics \& Statistics}, 7(1):1750861.

\bibitem[{Niu et~al.(2021)Niu, Tang, Zhang, Lu, Hua, and Wen}]{Niu_2021_CVPR}
Yulei Niu, Kaihua Tang, Hanwang Zhang, Zhiwu Lu, Xian-Sheng Hua, and Ji-Rong Wen. 2021.
\newblock \href {https://doi.org/10.1109/CVPR46437.2021.01251} {Counterfactual {VQA}: A cause-effect look at language bias}.
\newblock In \emph{Proceedings of the IEEE/CVF Conference on Computer Vision and Pattern Recognition (CVPR)}, pages 12700--12710.

\bibitem[{Puli et~al.(2022{\natexlab{a}})Puli, Joshi, He, and Ranganath}]{puli2022nuisances}
Aahlad Puli, Nitish Joshi, He~He, and Rajesh Ranganath. 2022{\natexlab{a}}.
\newblock \href {https://doi.org/10.48550/arXiv.2210.01302} {Nuisances via negativa: Adjusting for spurious correlations via data augmentation}.
\newblock \emph{arXiv preprint arXiv:2210.01302v2}.

\bibitem[{Puli et~al.(2022{\natexlab{b}})Puli, Zhang, Oermann, and Ranganath}]{puli2022outofdistribution}
Aahlad~Manas Puli, Lily~H Zhang, Eric~Karl Oermann, and Rajesh Ranganath. 2022{\natexlab{b}}.
\newblock \href {https://openreview.net/forum?id=12RoR2o32T} {Out-of-distribution generalization in the presence of nuisance-induced spurious correlations}.
\newblock In \emph{International Conference on Learning Representations}.

\bibitem[{Ren and Xiong(2023)}]{ren-xiong-2023-huaslim}
Yuqi Ren and Deyi Xiong. 2023.
\newblock \href {https://aclanthology.org/2023.findings-acl.781} {{H}ua{SLIM}: Human attention motivated shortcut learning identification and mitigation for large language models}.
\newblock In \emph{Findings of the Association for Computational Linguistics: ACL 2023}, pages 12350--12365, Toronto, Canada. Association for Computational Linguistics.

\bibitem[{Sagawa et~al.(2020)Sagawa, Koh, Hashimoto, and Liang}]{sagawa2020distributionally}
Shiori Sagawa, Pang~Wei Koh, Tatsunori~B. Hashimoto, and Percy Liang. 2020.
\newblock \href {https://openreview.net/forum?id=ryxGuJrFvS} {Distributionally robust neural networks}.
\newblock In \emph{International Conference on Learning Representations}.

\bibitem[{Sanh et~al.(2021)Sanh, Wolf, Belinkov, and Rush}]{sanh2021learning}
Victor Sanh, Thomas Wolf, Yonatan Belinkov, and Alexander~M Rush. 2021.
\newblock \href {https://openreview.net/forum?id=Hf3qXoiNkR} {Learning from others' mistakes: Avoiding dataset biases without modeling them}.
\newblock In \emph{International Conference on Learning Representations}.

\bibitem[{Schuster et~al.(2019)Schuster, Shah, Yeo, Roberto Filizzola~Ortiz, Santus, and Barzilay}]{schuster-etal-2019-towards}
Tal Schuster, Darsh Shah, Yun Jie~Serene Yeo, Daniel Roberto Filizzola~Ortiz, Enrico Santus, and Regina Barzilay. 2019.
\newblock \href {https://doi.org/10.18653/v1/D19-1341} {Towards debiasing fact verification models}.
\newblock In \emph{Proceedings of the 2019 Conference on Empirical Methods in Natural Language Processing and the 9th International Joint Conference on Natural Language Processing (EMNLP-IJCNLP)}, pages 3419--3425, Hong Kong, China. Association for Computational Linguistics.

\bibitem[{Shazeer et~al.(2017)Shazeer, Mirhoseini, Maziarz, Davis, Le, Hinton, and Dean}]{shazeer2017outrageously}
Noam Shazeer, Azalia Mirhoseini, Krzysztof Maziarz, Andy Davis, Quoc Le, Geoffrey Hinton, and Jeff Dean. 2017.
\newblock \href {https://openreview.net/forum?id=B1ckMDqlg} {Outrageously large neural networks: The sparsely-gated mixture-of-experts layer}.
\newblock In \emph{International Conference on Learning Representations}.

\bibitem[{Stacey et~al.(2022{\natexlab{a}})Stacey, Belinkov, and Rei}]{stacey2022supervising}
Joe Stacey, Yonatan Belinkov, and Marek Rei. 2022{\natexlab{a}}.
\newblock \href {https://doi.org/10.1609/aaai.v36i10.21386} {Supervising model attention with human explanations for robust natural language inference}.
\newblock \emph{Proceedings of the AAAI Conference on Artificial Intelligence}, 36(10):11349--11357.

\bibitem[{Stacey et~al.(2022{\natexlab{b}})Stacey, Minervini, Dubossarsky, and Rei}]{stacey-etal-2022-logical}
Joe Stacey, Pasquale Minervini, Haim Dubossarsky, and Marek Rei. 2022{\natexlab{b}}.
\newblock \href {https://aclanthology.org/2022.emnlp-main.251} {Logical reasoning with span-level predictions for interpretable and robust {NLI} models}.
\newblock In \emph{Proceedings of the 2022 Conference on Empirical Methods in Natural Language Processing}, pages 3809--3823, Abu Dhabi, United Arab Emirates. Association for Computational Linguistics.

\bibitem[{Stacey et~al.(2020)Stacey, Minervini, Dubossarsky, Riedel, and Rockt{\"a}schel}]{stacey-etal-2020-avoiding}
Joe Stacey, Pasquale Minervini, Haim Dubossarsky, Sebastian Riedel, and Tim Rockt{\"a}schel. 2020.
\newblock \href {https://doi.org/10.18653/v1/2020.emnlp-main.665} {{A}voiding the {H}ypothesis-{O}nly {B}ias in {N}atural {L}anguage {I}nference via {E}nsemble {A}dversarial {T}raining}.
\newblock In \emph{Proceedings of the 2020 Conference on Empirical Methods in Natural Language Processing (EMNLP)}, pages 8281--8291, Online. Association for Computational Linguistics.

\bibitem[{Thorne et~al.(2018)Thorne, Vlachos, Christodoulopoulos, and Mittal}]{thorne-etal-2018-fever}
James Thorne, Andreas Vlachos, Christos Christodoulopoulos, and Arpit Mittal. 2018.
\newblock \href {https://doi.org/10.18653/v1/N18-1074} {{FEVER}: a large-scale dataset for fact extraction and {VER}ification}.
\newblock In \emph{Proceedings of the 2018 Conference of the North {A}merican Chapter of the Association for Computational Linguistics: Human Language Technologies, Volume 1 (Long Papers)}, pages 809--819, New Orleans, Louisiana. Association for Computational Linguistics.

\bibitem[{Tian et~al.(2022)Tian, Cao, Zhang, and Xing}]{tian2022debiasing}
Bing Tian, Yixin Cao, Yong Zhang, and Chunxiao Xing. 2022.
\newblock \href {https://doi.org/10.1609/aaai.v36i10.21389} {Debiasing {NLU} models via causal intervention and counterfactual reasoning}.
\newblock \emph{Proceedings of the AAAI Conference on Artificial Intelligence}, 36(10):11376--11384.

\bibitem[{Titterington et~al.(1985)Titterington, Smith, and Makov}]{titterington1985statistical}
Donald~M. Titterington, Adrian~F.M. Smith, and Ehud Makov. 1985.
\newblock \href {https://books.google.co.jp/books?id=hZ0QAQAAIAAJ} {\emph{Statistical Analysis of Finite Mixture Distributions}}.
\newblock Applied section. Wiley.

\bibitem[{Utama et~al.(2020{\natexlab{a}})Utama, Moosavi, and Gurevych}]{utama-etal-2020-mind}
Prasetya~Ajie Utama, Nafise~Sadat Moosavi, and Iryna Gurevych. 2020{\natexlab{a}}.
\newblock \href {https://doi.org/10.18653/v1/2020.acl-main.770} {Mind the trade-off: Debiasing {NLU} models without degrading the in-distribution performance}.
\newblock In \emph{Proceedings of the 58th Annual Meeting of the Association for Computational Linguistics}, pages 8717--8729, Online. Association for Computational Linguistics.

\bibitem[{Utama et~al.(2020{\natexlab{b}})Utama, Moosavi, and Gurevych}]{utama-etal-2020-towards}
Prasetya~Ajie Utama, Nafise~Sadat Moosavi, and Iryna Gurevych. 2020{\natexlab{b}}.
\newblock \href {https://doi.org/10.18653/v1/2020.emnlp-main.613} {Towards debiasing {NLU} models from unknown biases}.
\newblock In \emph{Proceedings of the 2020 Conference on Empirical Methods in Natural Language Processing (EMNLP)}, pages 7597--7610, Online. Association for Computational Linguistics.

\bibitem[{Veitch et~al.(2021)Veitch, D\textquotesingle~Amour, Yadlowsky, and Eisenstein}]{veitch2021counterfactual}
Victor Veitch, Alexander D\textquotesingle~Amour, Steve Yadlowsky, and Jacob Eisenstein. 2021.
\newblock \href {https://proceedings.neurips.cc/paper_files/paper/2021/file/8710ef761bbb29a6f9d12e4ef8e4379c-Paper.pdf} {Counterfactual invariance to spurious correlations in text classification}.
\newblock In \emph{Advances in Neural Information Processing Systems}, volume~34, pages 16196--16208. Curran Associates, Inc.

\bibitem[{Wald(1950)}]{Wald1950WALSDF}
Abraham Wald. 1950.
\newblock \emph{Statistical Decision Functions}.
\newblock Wiley: New York.

\bibitem[{Walker and Ben-Akiva(2011)}]{walker2011advances}
Joan~L. Walker and Moshe Ben-Akiva. 2011.
\newblock \href {https://doi.org/10.4337/9780857930873.00015} {\emph{Advances in Discrete Choice: Mixture Models}}, chapter~8. Edward Elgar Publishing, Cheltenham, UK.

\bibitem[{Wang et~al.(2022)Wang, Xu, Szekely, and Chen}]{wang-etal-2022-robust}
Fei Wang, Zhewei Xu, Pedro Szekely, and Muhao Chen. 2022.
\newblock \href {https://doi.org/10.18653/v1/2022.naacl-main.371} {Robust (controlled) table-to-text generation with structure-aware equivariance learning}.
\newblock In \emph{Proceedings of the 2022 Conference of the North American Chapter of the Association for Computational Linguistics: Human Language Technologies}, pages 5037--5048, Seattle, United States. Association for Computational Linguistics.

\bibitem[{Wang and Culotta(2020)}]{wang-culotta-2020-identifying}
Zhao Wang and Aron Culotta. 2020.
\newblock \href {https://doi.org/10.18653/v1/2020.findings-emnlp.308} {Identifying spurious correlations for robust text classification}.
\newblock In \emph{Findings of the Association for Computational Linguistics: EMNLP 2020}, pages 3431--3440, Online. Association for Computational Linguistics.

\bibitem[{Wen et~al.(2022)Wen, Zhu, Zhang, Zhou, and Huang}]{wen-etal-2022-autocad}
Jiaxin Wen, Yeshuang Zhu, Jinchao Zhang, Jie Zhou, and Minlie Huang. 2022.
\newblock \href {https://aclanthology.org/2022.findings-emnlp.170} {{A}uto{CAD}: Automatically generate counterfactuals for mitigating shortcut learning}.
\newblock In \emph{Findings of the Association for Computational Linguistics: EMNLP 2022}, pages 2302--2317, Abu Dhabi, United Arab Emirates. Association for Computational Linguistics.

\bibitem[{Williams et~al.(2018)Williams, Nangia, and Bowman}]{williams-etal-2018-broad}
Adina Williams, Nikita Nangia, and Samuel Bowman. 2018.
\newblock \href {https://doi.org/10.18653/v1/N18-1101} {A broad-coverage challenge corpus for sentence understanding through inference}.
\newblock In \emph{Proceedings of the 2018 Conference of the North {A}merican Chapter of the Association for Computational Linguistics: Human Language Technologies, Volume 1 (Long Papers)}, pages 1112--1122, New Orleans, Louisiana. Association for Computational Linguistics.

\bibitem[{Wu and Gui(2022)}]{wu-gui-2022-less}
Ting Wu and Tao Gui. 2022.
\newblock \href {https://aclanthology.org/2022.coling-1.143} {Less is better: Recovering intended-feature subspace to robustify {NLU} models}.
\newblock In \emph{Proceedings of the 29th International Conference on Computational Linguistics}, pages 1666--1676, Gyeongju, Republic of Korea. International Committee on Computational Linguistics.

\bibitem[{Wu et~al.(2022)Wu, Gardner, Stenetorp, and Dasigi}]{wu-etal-2022-generating}
Yuxiang Wu, Matt Gardner, Pontus Stenetorp, and Pradeep Dasigi. 2022.
\newblock \href {https://doi.org/10.18653/v1/2022.acl-long.190} {Generating data to mitigate spurious correlations in natural language inference datasets}.
\newblock In \emph{Proceedings of the 60th Annual Meeting of the Association for Computational Linguistics (Volume 1: Long Papers)}, pages 2660--2676, Dublin, Ireland. Association for Computational Linguistics.

\bibitem[{Xiang et~al.(2019)Xiang, Yao, and Yang}]{xiang2019overview}
Sijia Xiang, Weixin Yao, and Guangren Yang. 2019.
\newblock \href {https://www.jstor.org/stable/26874187} {An overview of semiparametric extensions of finite mixture models}.
\newblock \emph{Statistical Science}, 34(3):pp. 391--404.

\bibitem[{Xiong et~al.(2021)Xiong, Chen, Pang, Cheng, Ma, and Lan}]{xiong2021uncertainty}
Ruibin Xiong, Yimeng Chen, Liang Pang, Xueqi Cheng, Zhi-Ming Ma, and Yanyan Lan. 2021.
\newblock \href {https://proceedings.neurips.cc/paper_files/paper/2021/file/71a8b2ffe0b594a5c1b3c28090384fd7-Paper.pdf} {Uncertainty calibration for ensemble-based debiasing methods}.
\newblock In \emph{Advances in Neural Information Processing Systems}, volume~34, pages 13657--13669. Curran Associates, Inc.

\bibitem[{Yaghoobzadeh et~al.(2021)Yaghoobzadeh, Mehri, Tachet~des Combes, Hazen, and Sordoni}]{yaghoobzadeh-etal-2021-increasing}
Yadollah Yaghoobzadeh, Soroush Mehri, Remi Tachet~des Combes, T.~J. Hazen, and Alessandro Sordoni. 2021.
\newblock \href {https://doi.org/10.18653/v1/2021.eacl-main.291} {Increasing robustness to spurious correlations using forgettable examples}.
\newblock In \emph{Proceedings of the 16th Conference of the European Chapter of the Association for Computational Linguistics: Main Volume}, pages 3319--3332, Online. Association for Computational Linguistics.

\bibitem[{Yang et~al.(2023)Yang, Zhang, Katabi, and Ghassemi}]{pmlr-v202-yang23s}
Yuzhe Yang, Haoran Zhang, Dina Katabi, and Marzyeh Ghassemi. 2023.
\newblock \href {https://proceedings.mlr.press/v202/yang23s.html} {Change is hard: A closer look at subpopulation shift}.
\newblock In \emph{Proceedings of the 40th International Conference on Machine Learning}, volume 202 of \emph{Proceedings of Machine Learning Research}, pages 39584--39622. PMLR.

\bibitem[{Yang et~al.(2018)Yang, Dai, Salakhutdinov, and Cohen}]{yang2018breaking}
Zhilin Yang, Zihang Dai, Ruslan Salakhutdinov, and William~W. Cohen. 2018.
\newblock \href {https://openreview.net/forum?id=HkwZSG-CZ} {Breaking the softmax bottleneck: A high-rank {RNN} language model}.
\newblock In \emph{International Conference on Learning Representations}.

\bibitem[{Yang et~al.(2019)Yang, Dai, Yang, Carbonell, Salakhutdinov, and Le}]{yang2019xlnet}
Zhilin Yang, Zihang Dai, Yiming Yang, Jaime Carbonell, Russ~R Salakhutdinov, and Quoc~V Le. 2019.
\newblock \href {https://proceedings.neurips.cc/paper_files/paper/2019/file/dc6a7e655d7e5840e66733e9ee67cc69-Paper.pdf} {Xlnet: Generalized autoregressive pretraining for language understanding}.
\newblock In \emph{Advances in Neural Information Processing Systems}, volume~32. Curran Associates, Inc.

\bibitem[{Yao et~al.(2022)Yao, Wang, Li, Zhang, Liang, Zou, and Finn}]{pmlr-v162-yao22b}
Huaxiu Yao, Yu~Wang, Sai Li, Linjun Zhang, Weixin Liang, James Zou, and Chelsea Finn. 2022.
\newblock \href {https://proceedings.mlr.press/v162/yao22b.html} {Improving out-of-distribution robustness via selective augmentation}.
\newblock In \emph{Proceedings of the 39th International Conference on Machine Learning}, volume 162 of \emph{Proceedings of Machine Learning Research}, pages 25407--25437. PMLR.

\bibitem[{Yu et~al.(2022)Yu, Jiang, Zhang, Niu, Sun, and Bing}]{yu-etal-2022-interventional}
Sicheng Yu, Jing Jiang, Hao Zhang, Yulei Niu, Qianru Sun, and Lidong Bing. 2022.
\newblock \href {https://aclanthology.org/2022.emnlp-main.799} {Interventional training for out-of-distribution natural language understanding}.
\newblock In \emph{Proceedings of the 2022 Conference on Empirical Methods in Natural Language Processing}, pages 11627--11638, Abu Dhabi, United Arab Emirates. Association for Computational Linguistics.

\bibitem[{Zhang et~al.(2022)Zhang, Ren, Wang, Wu, and Song}]{zhang-etal-2022-making}
Chen Zhang, Lei Ren, Jingang Wang, Wei Wu, and Dawei Song. 2022.
\newblock \href {https://aclanthology.org/2022.emnlp-main.217} {Making pretrained language models good long-tailed learners}.
\newblock In \emph{Proceedings of the 2022 Conference on Empirical Methods in Natural Language Processing}, pages 3298--3312, Abu Dhabi, United Arab Emirates. Association for Computational Linguistics.

\bibitem[{Zhang et~al.(2019)Zhang, Baldridge, and He}]{zhang-etal-2019-paws}
Yuan Zhang, Jason Baldridge, and Luheng He. 2019.
\newblock \href {https://doi.org/10.18653/v1/N19-1131} {{PAWS}: Paraphrase adversaries from word scrambling}.
\newblock In \emph{Proceedings of the 2019 Conference of the North {A}merican Chapter of the Association for Computational Linguistics: Human Language Technologies, Volume 1 (Long and Short Papers)}, pages 1298--1308, Minneapolis, Minnesota. Association for Computational Linguistics.

\end{thebibliography}
\bibliographystyle{acl_natbib}

\clearpage
\appendix

\section{Additional Figures}
\label{sec:appx figures}
\begin{figure}[t]
    \centering
    \includegraphics[width=0.90\columnwidth,keepaspectratio]{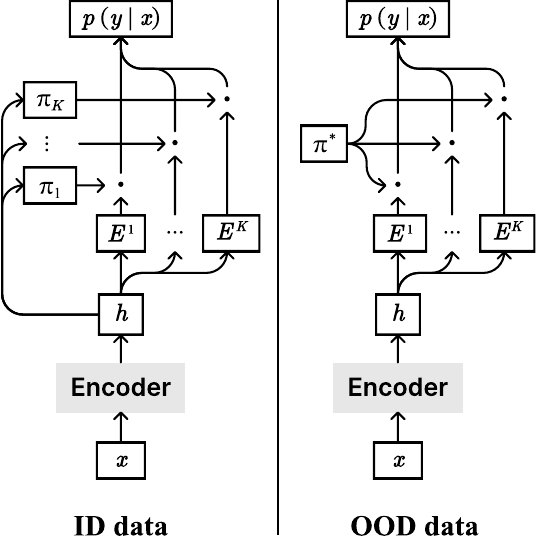}
    \caption{
    Overview of our method.
    We fit training data using a mixture model consisting of $K$ expert networks $\{E^k\}_{k=1}^K$ and a router network $\pi$ (Section~\ref{sec:mixture model}).
    During inference, the model is used as is for ID data, and $\pi$ is replaced with $\pi^*$ for OOD data (Section~\ref{sec:control}).
    }
    \label{fig:method overview}
\end{figure}

\begin{figure}[t]
    \centering    \includegraphics[width=0.72\columnwidth,keepaspectratio]{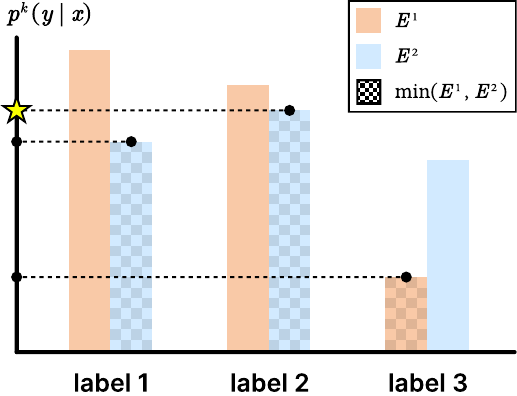}
    \caption{
    An example of decision-making with argmin weighting, where $K=2$ and $\mathcal{|Y|}=3$.
    After performing argmin weighting, label 2 achieves the highest score (starred) and is thus chosen as the answer.
    }
    \label{fig:argmin}
\end{figure}

Figure~\ref{fig:method overview} shows an overview of our method (Section~\ref{sec:method}), and Figure~\ref{fig:argmin} illustrates an example of our Argmin weighting (Section~\ref{sec:argmin}).

\begin{figure}[t]
    \centering
    \includegraphics[width=0.62\columnwidth,keepaspectratio]{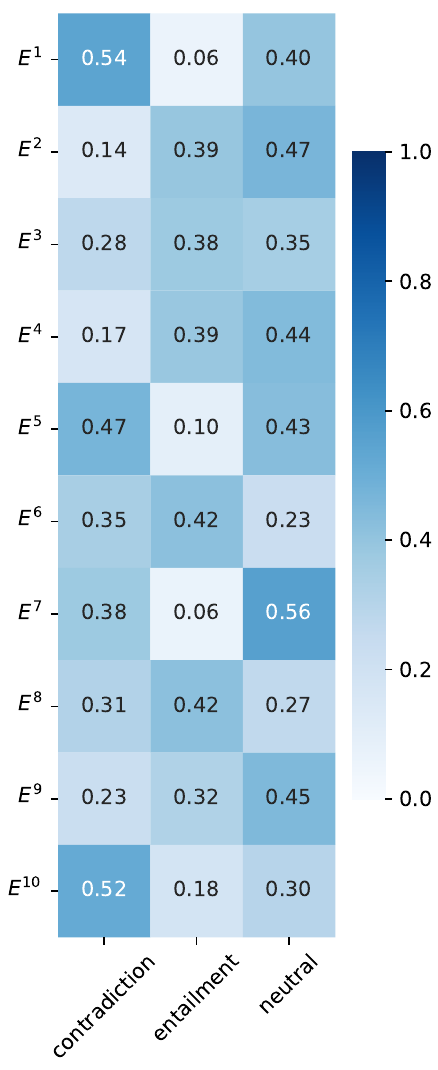}
    \caption{
    The average prediction of each expert $E^k$ across MNLI ID dev.
    }
    \label{fig:heatmap prediction}
\end{figure}

Figure~\ref{fig:heatmap prediction} shows the average prediction of each expert, calculated across MNLI ID dev (Section~\ref{sec:analysis penalty}).
We observe a significant variance between experts' predictions, which indicates that different experts tend to make different predictions.

\section{Further Setup Details}
\label{sec:appx setup}
Table~\ref{tab:setup details} specifies the URLs of the datasets, pre-trained models, and code of the previous methods we introduced in Section~\ref{sec:setup}.

Following the fine-tuning hyperparameters of DeBERTa$_{\text{v3-large}}$ \citep{he2023debertav}, we set the learning rate to 5e-6 and used gradient clipping with the maximum gradient norm of 1.0 in the ablation study with DeBERTa$_{\text{v3-large}}$ (Section~\ref{sec:analysis ablation}).

\begin{table*}[t]
\centering
\begin{adjustbox}{max width=0.9\textwidth}
\begin{tabular}{ll}  
\toprule
\multicolumn{2}{c}{\textbf{Datasets}} \\
\midrule
MNLI & \url{https://cims.nyu.edu/~sbowman/multinli/} \\
HANS & \url{https://github.com/tommccoy1/hans} \\
QQP and PAWS & \url{https://github.com/google-research-datasets/paws} \\
FEVER and Symm. v1/v2 & \url{https://github.com/TalSchuster/FeverSymmetric} \\
\midrule
\multicolumn{2}{c}{\textbf{Pre-Trained Models}} \\
\midrule
BERT & \url{https://huggingface.co/bert-base-uncased} \\
DeBERTa$_{\text{v3-large}}$ & \url{https://huggingface.co/microsoft/deberta-v3-large} \\
\midrule
\multicolumn{2}{c}{\textbf{Code}} \\
\midrule
Conf-reg~$\spadesuit~_{\text{self-debias}}$$^*$ & \url{https://github.com/UKPLab/emnlp2020-debiasing-unknown} \\
JTT$^*$ & \url{https://github.com/YyzHarry/SubpopBench} \\
RISK & \url{https://github.com/CuteyThyme/RISK} \\
EIIL & \url{https://github.com/PluviophileYU/BAI} \\
BAI & \url{https://github.com/PluviophileYU/BAI} \\
GroupDRO$_{\text{label-group}}$$^*$ & \url{https://github.com/YyzHarry/SubpopBench} \\
ReWeightCRT$^*$ & \url{https://github.com/YyzHarry/SubpopBench} \\
\bottomrule
\end{tabular}
\end{adjustbox}
\caption{
URLs of the datasets, pre-trained models, and code of the previous methods we used in the experiments.
The methods with $^*$ needed modification on the codes to cover all the datasets we used.
}
\label{tab:setup details}
\end{table*}

\end{document}